\documentclass[10pt,twocolumn,letterpaper]{article}
\usepackage{amsmath}
\usepackage{booktabs}   
\usepackage{multirow}   
\usepackage{siunitx}    
\usepackage{adjustbox}   
\usepackage{caption}    
\usepackage{amssymb}
\usepackage{fancyhdr}

\usepackage{cvpr}              

\definecolor{cvprblue}{rgb}{0.21,0.49,0.74}
\definecolor{lightblue}{rgb}{0.9,0.95,1}
\definecolor{lavender}{RGB}{206, 162, 212}
\usepackage[pagebackref,breaklinks,colorlinks,allcolors=cvprblue]{hyperref}

\def\paperID{1277} 
\def\confName{CVPR}
\def\confYear{2025}

\title{Unbiased Video Scene Graph Generation via Visual and Semantic Dual Debiasing}

\author{Yanjun Li$^1$$^\dag$, Zhaoyang Li$^1$, Honghui Chen$^2$, Lizhi Xu$^1$ \vspace{1pt} \\
$^1$University of Science and Technology of China\\
$^2$College of Physics and Information Engineering, Fuzhou University\\
{\tt\small ballinli@mail.ustc.edu.cn, lizhaoyang@mail.ustc.edu.cn, chh5840996@gmail.com}\\
{\tt\small xulizhi@mail.ustc.edu.cn} \quad 
{\small $^\dag$ Corresponding Author}\\
}

\begin{document}
\setcounter{page}{1}  
\maketitle
\thispagestyle{plain}  
\pagestyle{plain}  
\begin{abstract}
Video Scene Graph Generation (VidSGG) aims to capture dynamic relationships among entities by sequentially analyzing video frames and integrating visual and semantic information. However, VidSGG is challenged by significant biases that skew predictions. To mitigate these biases, we propose a \textbf{VI}sual and \textbf{S}emantic \textbf{A}wareness (VISA) framework for unbiased VidSGG. VISA addresses visual bias through memory-enhanced temporal integration that enhances object representations and concurrently reduces semantic bias by iteratively integrating object features with comprehensive semantic information derived from triplet relationships. This visual-semantics dual debiasing approach results in more unbiased representations of complex scene dynamics. Extensive experiments demonstrate the effectiveness of our method, where VISA outperforms existing unbiased VidSGG approaches by a substantial margin (\eg, \textbf{+13.1\%} improvement in mR@20 and mR@50 for the SGCLS task under Semi Constraint).

\begin{figure}[t]
  \centering
  \includegraphics[width=0.48\textwidth, height=0.54\textwidth]{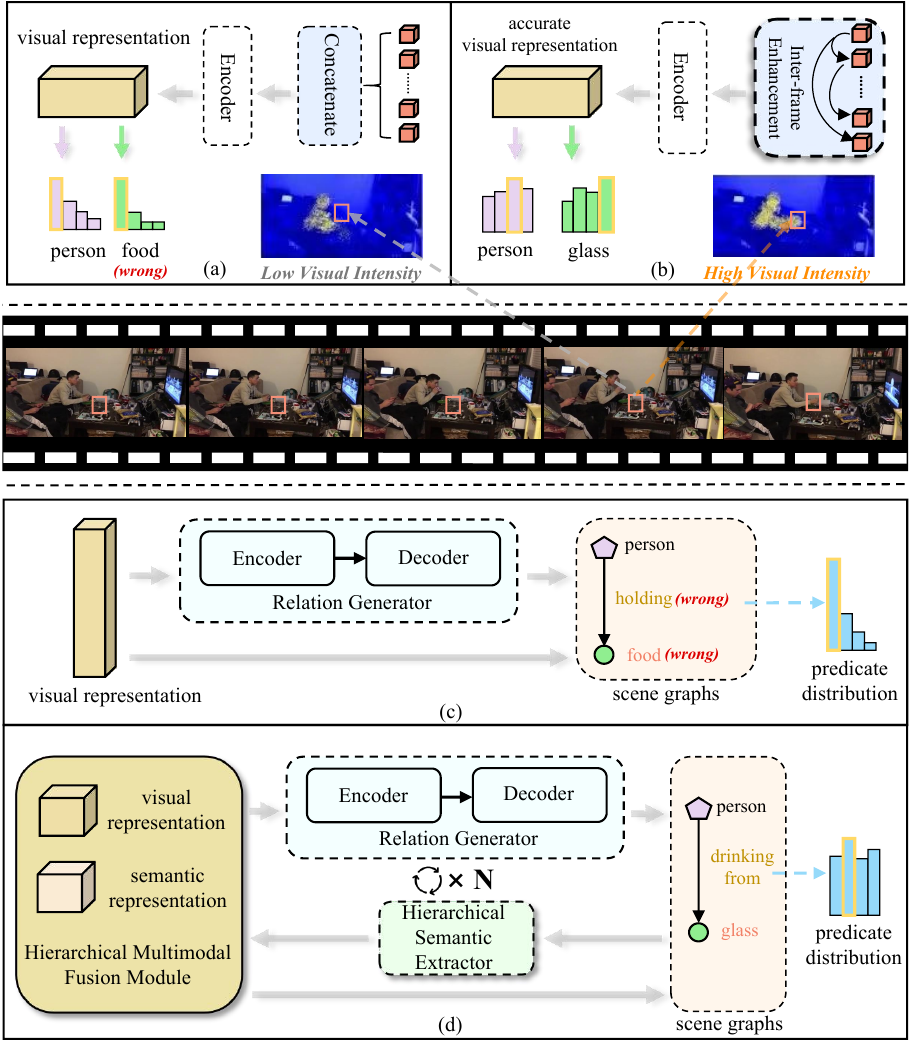}
  \caption{{\bf Framework Comparison.} We identify visual and semantic biases in VidSGG. Figures (a) and (c) show that prior methods rely on weak visual features and limited context, leading to inaccurate attention maps and skewed predicate distributions. In (b) and (d), our memory-enhanced visual debiasing and iterative semantic fusion mitigate both biases, resulting in stronger feature representations and more balanced predicate outcomes.}
  \vspace{-1em}
  \label{fig:motivation}
\end{figure}

\end{abstract}    
\vspace{-1.4em}
\section{Introduction}
\label{sec:intro}

Scene Graph Generation (SGG)~\cite{lu2016visual} constructs structured scene representations by identifying objects and their relationships as triplets (\texttt{<subject, predicate, object>}). This structure enables various downstream tasks~\cite{hildebrandt2020scene,li2022joint,schuster2015generating,gao2018image}, highlighting SGG as a pivotal area in computer vision. Initially applied to static images (ImgSGG)~\cite{xu2017scene}, SGG has been extended to video data (VidSGG)~\cite{chen2022video} to model dynamic interactions over time. However, VidSGG faces challenges due to biases favoring overrepresented classes, leading to the underrepresentation of rare classes.

Recent methods have adopted attention-based strategies to address class imbalance and mitigate classification biases~\cite{nag2023unbiased,pu2023spatial}. However, these approaches, primarily relying on learnable attention layers, often amplify biases towards high-frequency classes~\cite{khandelwal2024flocode}. To overcome this limitation, alternative methods have been proposed~\cite{li2021interventional,khandelwal2024flocode} aiming to reduce spurious correlations and better account for label correlations. Although these methods alleviate bias to some extent, they often fall short of achieving the desired performance. A key oversight of these approaches is their heavy reliance on debiasing techniques, which neglect the intrinsic visual-semantic nature of scene graphs.

Scene graphs inherently integrate both visual and semantic information, allowing biases to be categorized into \textbf{visual biases} and \textbf{semantic biases}. \textbf{Visual biases} arise from low visual quality in videos, such as blurring and occlusion~\cite{nag2023unbiased}, which degrade object feature representations. For instance, as shown in Fig.~\ref{fig:motivation}(a), objects may be partially visible or obscured over time, hindering models from capturing robust visual features for classification. Additionally, as illustrated in Fig.~\ref{fig:motivation}(c), the VidSGG task faces \textbf{semantic biases}. Previous methods~\cite{nag2023unbiased,cong2021spatial} rely solely on visual features to guide the relation generator for predicate prediction. This dependence on single-modal information leads to biased assumptions, similar to humans inferring relationships without sufficient context, thereby favoring more frequent predicate classes~\cite{liu2023cross}.

To address scene graph biases, we categorize them into visual and semantic types, guiding our proposed framework, \textbf{VISA} (\textbf{VI}sual and \textbf{S}emantic \textbf{A}wareness), for unbiased VidSGG, as illustrated in Fig.~\ref{fig:motivation}~(b) and~(d). VISA mitigates visual bias through a Memory-Enhanced Temporal Integrator that enriches the representations of entities in the current frame with historical representations from previous frames. It employs a relation generator to construct initial scene graphs and a hierarchical semantic extractor to iteratively refine semantic information. Our hierarchical fusion strategy integrates multimodal visual and semantic contexts to address semantic biases. Specifically, the visual debiasing method reduces visual feature variance, while the semantic debiasing method increases the Kullback-Leibler (KL) divergence. This dual-debiasing approach effectively mitigates \textbf{visual biases} caused by poor visual feature quality and \textbf{semantic biases} arising from insufficient contextual information, while also addressing model bias due to data skew. Extensive experiments demonstrate the effectiveness of VISA, achieving superior performance in unbiased VidSGG compared to previous methods. Our main contributions are summarized as follows:

\begin{itemize}
    \item We identify and explicitly define the visual and semantic biases affecting VidSGG, highlighting their impact on model performance.
    \item We propose VISA, a debiasing framework for VidSGG that simultaneously addresses visual and semantic biases through memory-enhanced temporal integration and hierarchical semantic extraction.
    \item We validate the effectiveness of our approach through extensive experiments and theoretical analyses, demonstrating state-of-the-art performance in unbiased VidSGG.
\end{itemize}

\vspace{-0.5em}
\section{Related Work}
\label{sec}

\subsection{Scene Graph Generation}
Scene Graph Generation (SGG) approaches focus on static image-based SGG (ImgSGG) and video-based SGG (VidSGG). For ImgSGG, sequential encoders like LSTM~\cite{zellers2018neural} and attention mechanisms~\cite{vaswani2017attention} capture global context. Recent works~\cite{li2022sgtr, shit2022relationformer, cong2023reltr} improve entity and predicate proposals, with SGTR~\cite{li2022sgtr} advancing this through a transformer-based architecture for more accurate scene graph generation.

VidSGG extends Scene Graph Generation (SGG) to dynamic contexts by incorporating intra-frame relationships, crucial for capturing temporal dependencies~\cite{ji2020action}. Recent advancements enhance temporal coherence through improved modeling of temporal dependencies~\cite{liu2020beyond}, while efficiency is boosted by adaptive structures~\cite{teng2021target}. Additionally, social context modeling~\cite{chen2021social} refines the understanding of interactions between multiple agents, further improving scene graph accuracy and robustness.

Despite their success, existing methods ignore unbiasedness. This results in biased SGG with inaccurate and misleading relationships in scenes. Consequently, such biases make models less reliable and less generalizable to real-world applications.

\subsection{Unbiased Scene Graph Generation}
Unlike conventional SGG, unbiased SGG focuses on reducing biases towards frequent objects and relationships. Key approaches in this area include knowledge distillation~\cite{li2023label} and dual-branch architectures~\cite{zheng2023dual}, which tackle different aspects of bias. For example, LS-KD~\cite{li2023label} uses knowledge distillation to address multi-predicate challenges by leveraging a teacher-student framework to improve relationship diversity. Building on this, DHL~\cite{zheng2023dual} employs a dual-branch architecture to ensure balanced attention between head and tail classes, preventing the dominance of common predicates. These methods complement each other by addressing bias from different angles, contributing to more accurate and balanced scene graph generation.

For unbiased VidSGG, Transformer-based models~\cite{cong2021spatial} and Gaussian Mixture Models~\cite{nag2023unbiased} have been explored to capture video dynamics better and reduce bias in visual relations. Iterative approaches~\cite{shang2021video} use conditional variables to improve video relation detection but lack a hierarchical strategy that integrates both visual and semantic representations, as in our method.

Unlike existing approaches that merely apply traditional debiasing strategies to VidSGG, our framework leverages the inherent visual-semantic nature of scene graphs. This enables a solution that is both experimentally validated and theoretically robust, effectively addressing the fundamental challenges in achieving unbiased VidSGG.

\vspace{-0.1em}
\section{Method}
\label{sec:Method}
\vspace{-0.2em}
\begin{figure*}[t]
  \centering
  \includegraphics[width=1\textwidth]{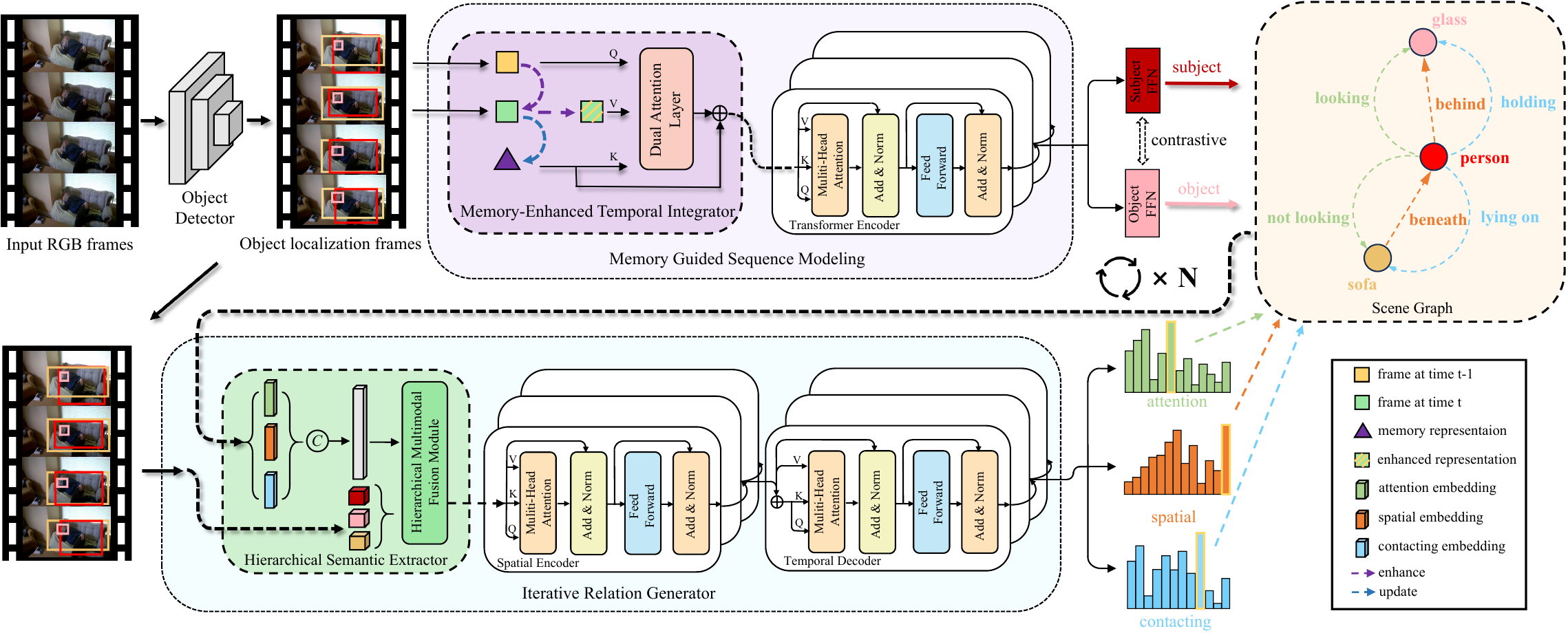}
  \caption{{\bf Framework of VISA.} 
   Given an input video, VISA extracts object features using an off-the-shelf detector. Subsequently, the Memory Guided Sequence Modeling leverages these features to develop robust object representations. Following this, the Iterative Relation Generator employs the object features to predict relationships between subject-object pairs. It is important to note that the Hierarchical Semantic Extractor is not engaged during the initial phase of scene graph generation.}
  \vspace{-1em}
  \label{fig:framework}
\end{figure*}
 
\subsection{Overview}
The VISA framework, depicted in Fig.~\ref{fig:framework}, comprises Memory Guided Sequence Modeling (MGSM), and the Iterative Relation Generator (IRG). It begins with an off-the-shelf object detector to identify objects in video frames, $\mathbf{b}_i^t$ represents the object anchors of the $i$-th entity in the $t$-th frame. In bounding-box approaches (e.g., VidSGG), $\mathbf{b}_i^t \in \mathbb{R}^4$ denotes a box, while in panoptic segmentation, $\mathbf{b}_i^t \in \mathbb{R}^n$ represents the $n$ boundary points. RoI-Align operations~\cite{he2017mask} are then applied to obtain feature representations, denoted as \( \mathbf{v}_{i}^{t} \). The MGSM refines these features to produce subject and object representations, yielding enhanced features \( \mathbf{\hat{v}}_i^{t} \), while two separate FFNs generate distinct subject and object embeddings. The IRG leverages these visual representations, using the Hierarchical Semantic Extractor (HSE) to fuse visual and semantic information. Through iterative processing, it deduces unbiased semantic relations and generates relation predictions.

\subsection{Memory Guided Sequence Modeling}
\label{sec:MGSM}

We observe that as video duration increases, the Transformer tends to generate insufficient visual representations, particularly for small objects prone to blurring and occlusion (see Fig.~\ref{fig:motivation}~(a)). This insufficiency leads to high variance in feature representations due to the instability of visual information, causing the model to favor frequent entities and overlook less common ones.

Our theoretical analysis reveals that this instability introduces noise into the object representations, modeled as: \begin{equation} \mathbf{v}_i^t = \mathbf{v}_i + \boldsymbol{\epsilon}_i^t, \end{equation} where $\boldsymbol{\epsilon}_i^t$ denotes zero-mean Gaussian noise with covariance $\boldsymbol{\Sigma}$, \ie, $\mathbb{E}[\boldsymbol{\epsilon}_i^t] = \mathbf{0}$ and $\operatorname{Cov}[\boldsymbol{\epsilon}_i^t] = \boldsymbol{\Sigma}$. We assume these noise terms are independent across different time steps and objects. The observed features from the $t$-th frame are used directly as the object's representation: \begin{equation} \hat{\mathbf{v}}_i^t = \mathbf{v}_i^t. \end{equation}

Consequently, the expectation and variance of the visual representation are computed as follows:
\begin{equation}
\begin{aligned}
\mathbb{E}[\hat{\mathbf{v}}_i^t] &= \mathbf{v}_i,\quad \operatorname{Var}[\hat{\mathbf{v}}_i^t] = \boldsymbol{\Sigma}.
\end{aligned}
\label{eq:expectation_variance}
\vspace{-0.5em}
\end{equation}

While the single-frame estimation is unbiased, the noise introduced due to visual instability leads to high variance in the feature representations. This high variance destabilizes the feature representations and impairs subsequent relational predictions.

\begin{figure}[t]
  \centering
  \includegraphics[width=0.45\textwidth]{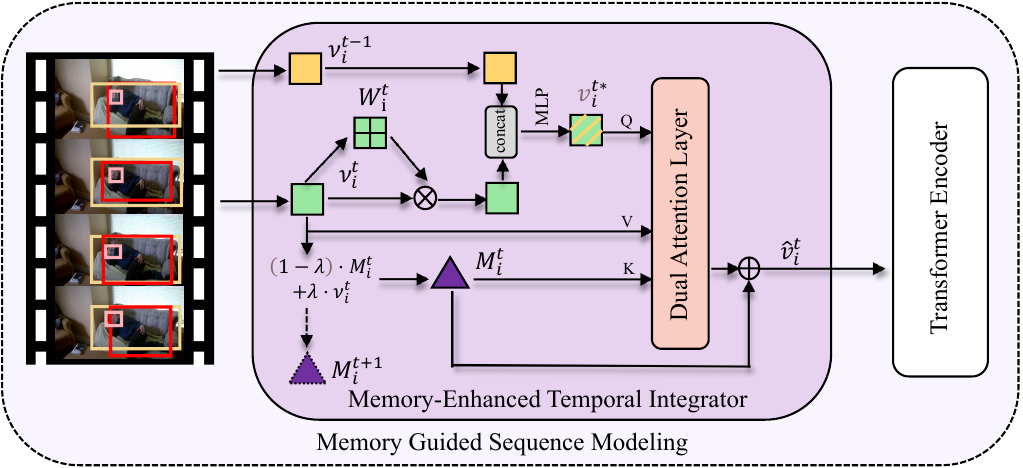}
  \caption{{\bf Structure of Memory Guided Sequence Modeling.} }
  \vspace{-1em}
  \label{fig:MGSM}
\end{figure}

Understanding that the high variance stems from the instability of visual information motivated us to design the \textbf{Memory Guided Sequence Modeling (MGSM)} module (illustrated in Fig.~\ref{fig:MGSM}). The MGSM leverages temporal visual information from preceding frames to enhance individual feature representations of frames, effectively reducing the variance caused by insufficient visual representations.

Specifically, we compute adaptive weights $\mathbf{W}_i^t$ based on the current frame's features: \begin{equation} \mathbf{W}_i^t = \sigma(\mathrm{MLP}(\mathbf{v}_i^t)), \end{equation} where $\sigma$ is the sigmoid function, and $\mathrm{MLP}$ denotes a multi-layer perceptron. Using these weights, we obtain the weighted feature representation: \begin{equation} \mathbf{v}_i^{t*} = \mathbf{W}_i^t \odot \mathbf{v}_i^t \Vert \mathbf{v}_i^{t-1}, \end{equation} where $\odot$ denotes element-wise multiplication, and $\Vert$ indicates feature concatenation.

To accumulate visual information and refine the current representations, we introduce an update mechanism:
\begin{equation} \mathbf{M}_i^{t+1} = (1 - \lambda) \mathbf{M}_i^t + \lambda \mathbf{v}_i^t. \end{equation} 
As a result, the expectation and variance of the memory representation $\mathbf{M}_i^t$ stabilize over time: \begin{equation} 
\mathbb{E}[\mathbf{M}_i^t] = \mathbf{v}_i, \operatorname{Var}[\mathbf{M}_i^t] = \frac{\lambda \boldsymbol{\Sigma}}{2 - \lambda} \approx \frac{\lambda \boldsymbol{\Sigma}}{2}, \text{for small } \lambda.
\end{equation}  This analysis shows that a smaller $\lambda$ reduces variance, enhancing the robustness of feature estimation. However, $\lambda$ cannot be too small, as excessively small values slow the adaptation to new information and may introduce bias. We provide detailed proofs of the variance calculation and the associated trade-off in the supplementary materials.

The final enhanced feature representation is obtained through a dual attention layer: \begin{equation} \hat{\mathbf{v}}_i^t = \mathrm{Attention}(\mathbf{Q} = \mathbf{v}_i^{t*}, \mathbf{K} = \mathbf{M}_i^t, \mathbf{V} = \mathbf{v}_i^t). \end{equation} Finally, the final enhanced feature representation is processed by the Transformer Encoder, as in~\cite{nag2023unbiased}, to generate the subject and object representations. By adaptively enhancing features and leveraging the reduced variance from the memory representation, the MGSM addresses the high variance caused by insufficient visual representations, thereby mitigating visual bias.

Our experimental results (Fig.~\ref{fig:visual}) confirm our theoretical insights by demonstrating that MGSM enhances visual representations. This enhancement reduces the high variance caused by insufficient visual information, which is crucial for addressing visual bias in VidSGG.

\subsection{Iterative Relation Generator}
\label{sec:IRG}

Existing methods for unbiased SGG predominantly predict predicates directly from visual representations~\cite{cong2021spatial,nag2023unbiased}. However, this approach is akin to humans attempting to deduce relationships solely based on visual cues without sufficient contextual information, which easily introduces semantic bias~\cite{liu2023cross}. Specifically, when lacking context, models tend to over-rely on the imbalanced and biased prior distributions present in the training data, leading to biased relationship predictions.

To alleviate semantic bias, we start by analyzing the problem from an information-theoretic perspective. Semantic bias occurs when a model lacks sufficient contextual information, leading to high uncertainty in predicting relationships. This uncertainty causes the model to over-rely on biased prior distributions from the training data. We use entropy $H$ to quantify this uncertainty~\cite{shannon1948mathematical}. We consider the conditional entropy $H(r_{ij} \mid v_i, v_j, S)$, where $v_i$ and $v_j$ are the feature representations of the subject and object, respectively. Based on the properties of conditional entropy~\cite{shannon1948mathematical}, the additional context provided by $S$ decreases the uncertainty in predicting relationships:
\vspace{-0.2em}
\begin{equation}
H(r_{ij} \mid v_i, v_j, S) \leq H(r_{ij} \mid v_i, v_j).
\end{equation}

This reduction in conditional entropy means the model is less uncertain and relies less on biased priors. By incorporating context $S$, the likelihood term $P(v_i, v_j, S \mid r_{ij})$ in Bayes' theorem becomes more informative, effectively reducing the relative influence of the prior $P(r_{ij})$. According to Bayes' theorem:

\vspace{-1em}
\begin{equation}
P(r_{ij} \mid v_i, v_j, S) = \frac{P(v_i, v_j, S \mid r_{ij}) \cdot P(r_{ij})}{P(v_i, v_j, S)}.
\end{equation}

To quantify the change in reliance on the prior, we consider the Kullback-Leibler (KL) divergence between the posterior $P(r_{ij} \mid v_i, v_j, S)$ and the prior $P(r_{ij})$:

\vspace{-1.5em}
\begin{align}
D_{\text{KL}}(&P(r_{ij} \mid v_i, v_j, S) \| P(r_{ij})) \nonumber \\
&= \sum_{r_{ij}} P(r_{ij} \mid v_i, v_j, S) \log \frac{P(r_{ij} \mid v_i, v_j, S)}{P(r_{ij})} \nonumber \\
&= H\left(r_{ij} \mid v_i, v_j, P(r_{ij})\right) - H\left(r_{ij} \mid v_i, v_j, S\right).
\end{align}
\vspace{-1em}

As the conditional entropy $H(r_{ij} \mid v_i, v_j, S)$ decreases due to the additional context, the KL divergence increases. This indicates that the posterior distribution diverges from the biased prior, reflecting a reduced reliance on the prior and effectively mitigating semantic bias.

This theoretical insight motivates our approach to incorporate contextual information into the model. Based on this, we propose the \textbf{Iterative Relation Generator (IRG)}, an iterative semantic debiasing component within our framework. The IRG iteratively introduces contextual information to refine relationship predictions, thereby reducing uncertainty and dependence on biased priors.

In the first iteration, the IRG infers semantic information from each subject-object pair to generate semantic embeddings. For each subject-object pair $(j, i)$ at frame $t$, we construct the composite objects features $\mathbf{p}_{j,i}^{t}$:

\vspace{-1.5em}
\begin{equation}
\mathbf{p}_{j,i}^{t} = f_v(\mathbf{v}_{j}^{t}) \Vert f_v(\mathbf{v}_{i}^{t}) \Vert \left( f_u(\mathbf{u}_{ji}^{t}) + f_{\text{box}}(b_{j}^{t}, b_{i}^{t}) \right) \Vert \mathbf{s}_j^{t} \Vert \mathbf{s}_i^{t},
\vspace{0.5em}
\end{equation}

\noindent where $\mathbf{v}_{j}^{t}$ and $\mathbf{v}_{i}^{t}$ are the visual feature representations of the subject and object. The tensor $\mathbf{u}_{ji}^{t}$ is the union box feature map extracted via RoIAlign~\cite{he2017mask}, and $\mathbf{s}_j^{t}$ and $\mathbf{s}_i^{t}$ are the semantic embeddings obtained from GloVe~\cite{pennington2014glove}. The functions $f_v$, $f_u$, and $f_{\text{box}}$ are feature transformation functions.

We then perform initial relationship prediction, obtaining predicate categories such as \textit{attention}, \textit{spatial}, and \textit{contacting}. We select the semantic embeddings with the highest probabilities from these distributions to form integrated triplet embeddings $\mathbf{C}_{pre,(j,i)}^t$:
\vspace{-0.2em}
\begin{equation}
\mathbf{C}_{pre,(j,i)}^t = \left[
\begin{array}{c}
f_v(\mathbf{v}_{j}^{t}) \Vert f_{a,(j,i)}^{t} \Vert f_v(\mathbf{v}_{i}^{t}) \\
f_v(\mathbf{v}_{j}^{t}) \Vert f_{s,(j,i)}^{t} \Vert f_v(\mathbf{v}_{i}^{t}) \\
f_v(\mathbf{v}_{j}^{t}) \Vert f_{c,(j,i)}^{t} \Vert f_v(\mathbf{v}_{i}^{t})
\end{array}
\right],
\end{equation}

\noindent where $\scalebox{0.9}{$f_{a,(j,i)}^{t}$}$, $\scalebox{0.9}{$f_{s,(j,i)}^{t}$}$, and $\scalebox{0.9}{$f_{c,(j,i)}^{t}$}$ are the semantic embeddings corresponding to the respective predicates.

\begin{figure}[t]
  \centering
  \includegraphics[width=0.48\textwidth]{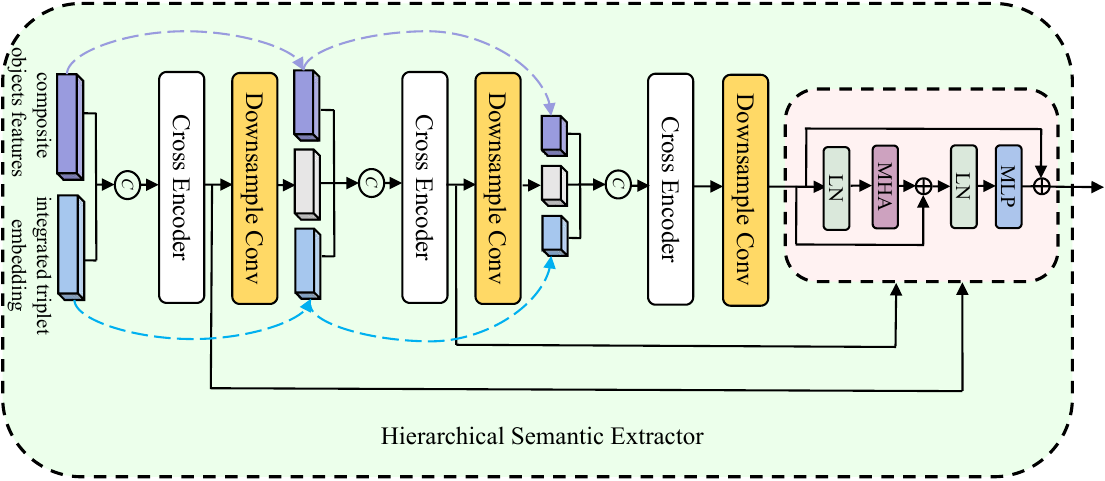}
  \caption{{\bf Structure of Hierarchical Semantics Extractor.} }
  \vspace{-1em}
  \label{fig:HSE}
\end{figure}

To further exploit hierarchical contextual information, we design the \textbf{Hierarchical Semantics Extractor (HSE)} (as shown in Fig.~\ref{fig:HSE}). Inspired by the multi-hierarchical feature fusion method~\cite{wang2023unified}, the HSE combines composite object features with integrated triplets to obtain fine-grained visual and semantic representations. We decompose the composite objects features $\mathbf{p}_{j,i}^{t}$ into fine-grained subject and object representations as follows:
\vspace{-0.1em}
\begin{equation}
\begin{aligned}
\mathbf{S}_{\mathbf{p}_{j,i}^{t}}^t &= f_v(\mathbf{v}_{j}^{t}) \Vert \left( f_u(\mathbf{u}_{ji}^{t}) + f_{\text{box}}(b_{j}^{t}, b_{i}^{t}) \right) \Vert \mathbf{s}_j^{t}, \\
\mathbf{O}_{\mathbf{p}_{j,i}^{t}}^t &= f_v(\mathbf{v}_{i}^{t}) \Vert \left( f_u(\mathbf{u}_{ji}^{t}) + f_{\text{box}}(b_{j}^{t}, b_{i}^{t}) \right) \Vert \mathbf{s}_i^{t}.
\end{aligned}
\end{equation}

\noindent Similarly, we decompose the integrated triplet embeddings $\mathbf{C}_{pre,(j,i)}^t$:

\begin{equation}
\begin{aligned}
\mathbf{S}_{\mathbf{C}_{(j,i)}^t}^t &= \left\{ f_v(\mathbf{v}_{j}^{t}) \Vert f_k^{t} \mid k \in \{ a, s, c \} \right\}, \\
\mathbf{O}_{\mathbf{C}_{(j,i)}^t}^t &= \left\{ f_k^{t} \Vert f_v(\mathbf{v}_{i}^{t}) \mid k \in \{ a, s, c \} \right\}.
\end{aligned}
\end{equation}

These fine-grained representations are concatenated with downsampled embeddings from the previous cross-attention layer via a stride-two convolutional layer, serving as inputs to the next cross-attention layer. This hierarchical structure enables the model to capture multi-level contextual information, effectively reducing uncertainty and reliance on biased priors. The HSE is active for iteration counts $N \geq 1$.

After $N$ iterations of the IRG, the final inter-object predicates are generated by the Spatial Encoder and Temporal Decoder modules adopted from~\cite{cong2021spatial}. By iteratively leveraging contextual information, our method effectively mitigates semantic bias. The experimental results (as shown in Fig.~\ref{fig:matrix}) validate our theoretical insights by demonstrating that incorporating contextual information reduces the model's reliance on biased priors and improves relationship predictions.

\subsection{Training and Testing}
\noindent\textbf{Training.}~In the training phase, VISA synthesizes object features from MGSM for preliminary scene graph construction. Upon reaching iteration \(N \geq 1\), it activates the HSE to generate a hierarchical visual-semantics fusion embedding. In Section~\ref{sec:MGSM}, the memory representation $\mathbf{M}_i^t$ is initialized as a zero tensor, while the embeddings for the three relationships are randomly initialized as described in Section~\ref{sec:IRG}. We optimize the comprehensive loss function:

\vspace{-0.5em}
\begin{equation}
    L_{\text{total}} = L_p + L_e + L_{\text{contra}}. \
\end{equation}
\vspace{-1em}

Here, \( L_p \) and \( L_e \) denote the predicate and entity classification losses, both computed via cross-entropy, and \( L_{\text{contra}} \) is the contrastive loss, all defined as in~\cite{nag2023unbiased}.

\noindent\textbf{Testing.}~After training, the proposed MGSM produces object predictions through the visual debiasing component, while the IRG performs semantic debiasing.

\begin{figure*}[t!]
  \vspace{-1em}
  \centering
  \includegraphics[width=0.9\textwidth]{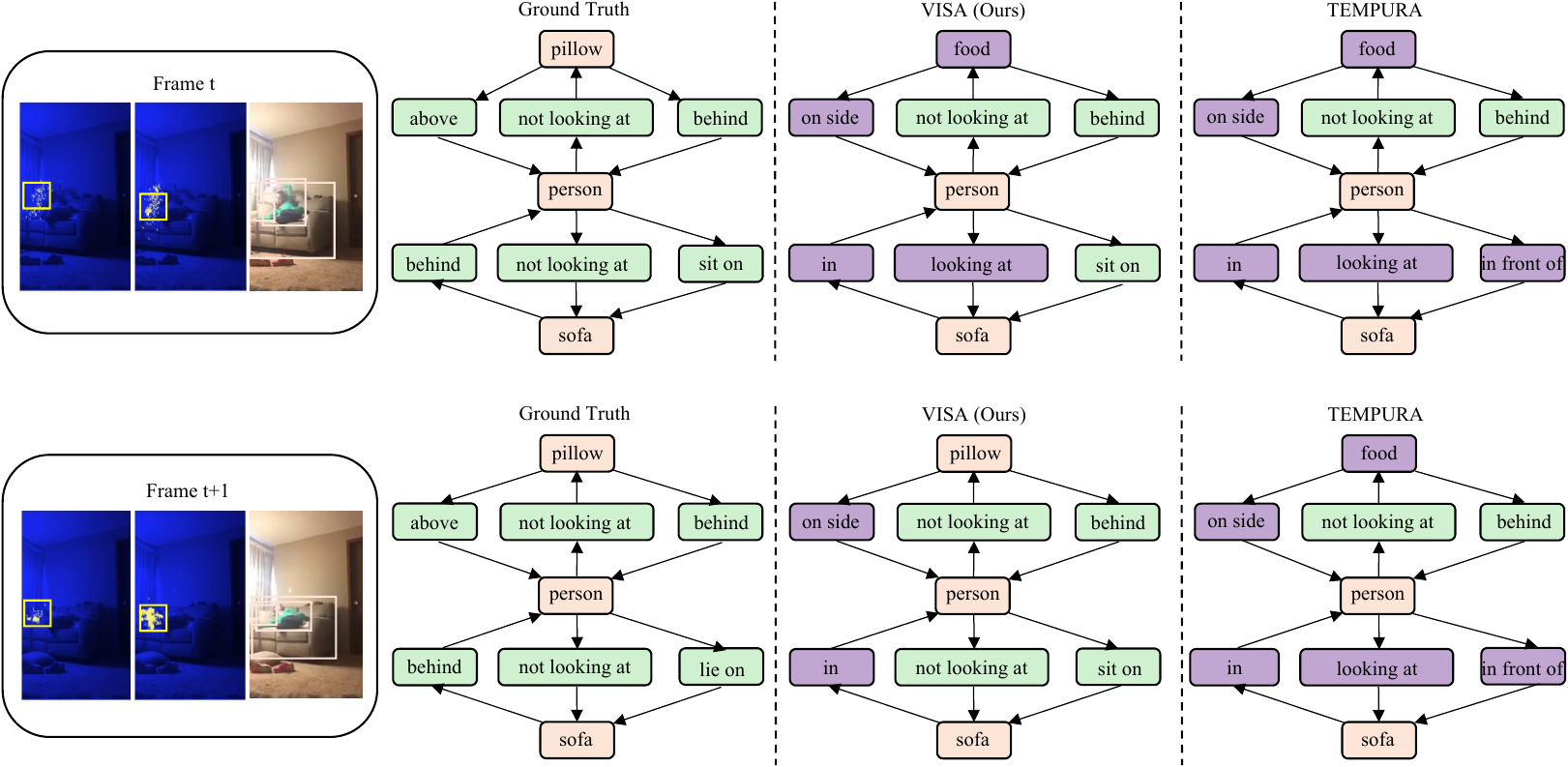}
  \vspace{0em}
  \caption{{\bf Qualitative results of our proposed method.} Comparison with the open-sourced state-of-the-art model TEMPURA~\cite{nag2023unbiased} under Semi Constraint, closely matching ground truth~\cite{cong2021spatial}. Visualizations show attention intensity and video scene graphs from Frame \( t \) to \( t+1 \). False positive relationships and object classifications are highlighted in \textbf{\textcolor{lavender}{purple}} blocks.}
  \label{fig:visualization} 
\end{figure*}

\begin{figure}[t]
  \centering
    \includegraphics[width=0.5\textwidth,height=0.35\textwidth,keepaspectratio]{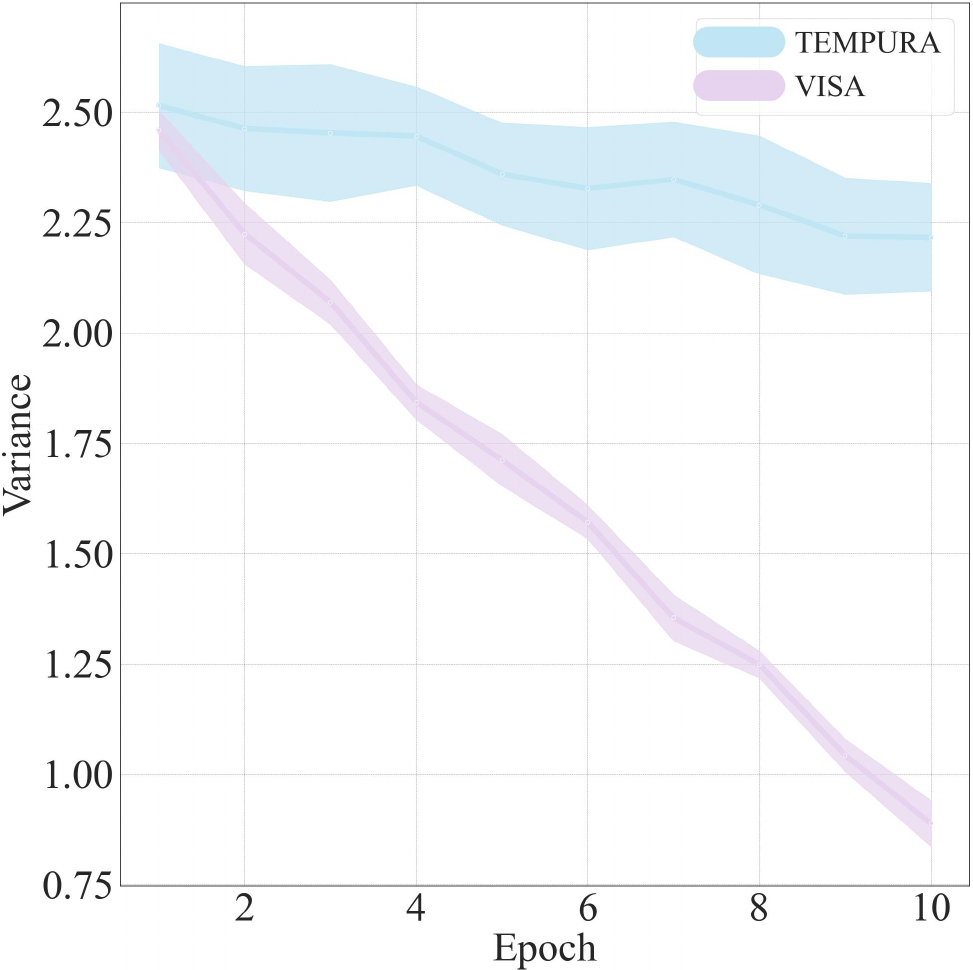}
    \vspace{-0.5em}
  \caption{{\bf Comparison of TEMPURA and VISA visual feature representation variance.} Visualization of visual feature representation variance across training epochs. The shaded regions represent the standard deviation. }

  \label{fig:visual}
  \vspace{-0.5em}
\end{figure}

\section{Experiments}
\label{sec:Experiments}
\subsection{Dataset and Metrics}
\noindent\textbf{Dataset.}~We conducted experiments on the Action Genome (AG) dataset~\cite{ji2020action}, the largest benchmark for VidSGG. AG consists of $234,253$ annotated frames containing $476,229$ bounding boxes across $35$ object categories. It provides $1,715,568$ predicate instances spanning $26$ distinct relationship classes.
\vspace{0.2em}

\noindent\textbf{Metrics and Evaluation Setup.}~We employed the standard metrics for unbiased VidSGG evaluation, mean Recall@K (mR@K), considering $K \in \{10, 20, 50\}$. Following established protocols~\cite{ji2020action,cong2021spatial,nag2023unbiased}, we engaged in three SGG tasks, PREDCLS, SGCLS and SGDET. Evaluation was conducted across three settings, With Constraint, Semi Constraint, and No Constraint. Due to page constraints, further details are provided in the supplementary materials.

\begin{figure}[t!]
  \centering
  \includegraphics[width=0.48\textwidth]{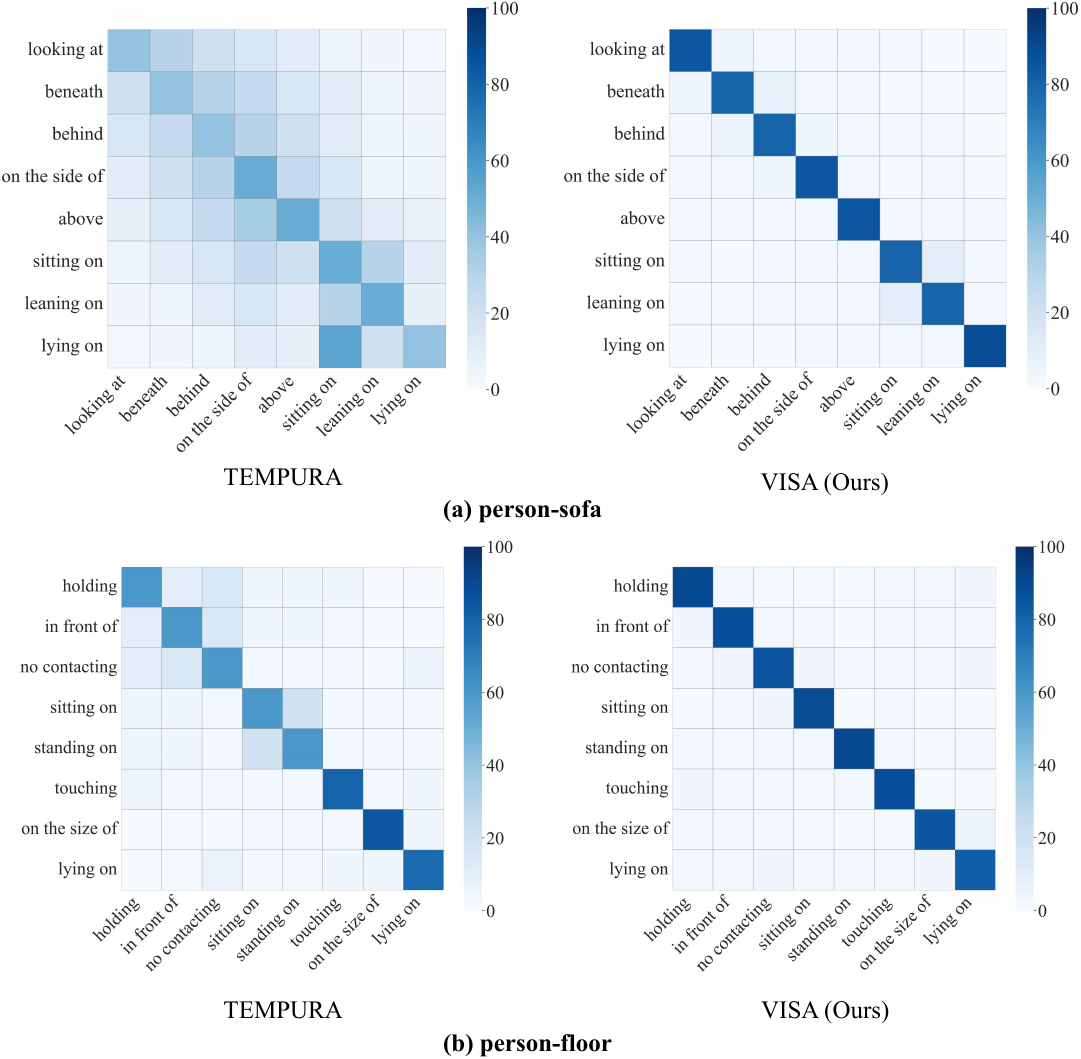}
  \vspace{-1em}
  \caption{{\bf Qualitative results for relationship confusion matrix.} Results for the subject-object pairs, person-sofa, and person-floor.}
  \vspace{-2em}
  \label{fig:matrix} 
\end{figure}

\subsection{Implementation details}
Following prior work~\cite{cong2021spatial,li2022dynamic,nag2023unbiased}, we adopted Faster R-CNN~\cite{ren2015faster} with ResNet-101~\cite{he2016deep} as the object detector, initially trained on the AG dataset. To ensure a fair comparison, we utilized the official implementations of these methods. For our MGSM module, we set the $\lambda$ parameter to $0.04$ for the \textit{SGCLS} task and $0.06$ for the \textit{SGDET} task. In the IRG module, we implemented a dual-procedure setup, enabling iterative relational inference with the number of iterations $N$ set to $1$. Due to page constraints, further details are provided in the supplementary materials.

\begin{table*}[t]
\centering
\caption{Quantitative results comparisons with unbiased SGG methods under With, Semi, and No Constraints. The experiments are conducted on the dataset AG~\cite{ji2020action}. The \textbf{best} results are highlighted.}
\scriptsize
\setlength{\tabcolsep}{3pt}
\renewcommand{\arraystretch}{1}
\begin{tabular}{l l c c c c c c c c c}
\toprule
\multirow{2}{*}{Constraint} & \multirow{2}{*}{Method} & 
\multicolumn{3}{c}{PREDCLS} & 
\multicolumn{3}{c}{SGCLS} &  
\multicolumn{3}{c}{SGDET}  \\ 
\cmidrule(lr){3-5} \cmidrule(lr){6-8} \cmidrule(lr){9-11} 
& & mR@10 & mR@20 & mR@50 & mR@10 & mR@20 & mR@50 & mR@10 & mR@20 & mR@50 \\
\midrule

\multirow{10}{*}{\rotatebox{90}{With}} 
& RelDN~\cite{zhang2019graphical} & 6.2 & 6.2 & 6.2 & 3.4 & 3.4 & 3.4 & 3.3 & 3.3 & 3.3 \\
& HCRD supervised \cite{carreira2017quo} & - & - & - & - & - & - & - & 8.3 & 9.1 \\
& TRACE~\cite{teng2021target} & 15.2 & 15.2 & 15.2 & 8.9 & 8.9 & 8.9 & 8.2 & 8.2 & 8.2 \\
& ISGG \cite{khandelwal2022iterative} & - & - & - & - & - & - & 19.7 & 22.9 & - \\
& STTran~\cite{cong2021spatial} & 37.8 & 40.1 & 40.2 & 27.2 & 28.0 & 28.0 & 20.8 & 20.8 & 22.2 \\ 
& STTran-TPI~\cite{wang2022dynamic} & 37.3 & 40.6 & 40.6 & 28.3 & 29.3 & 29.3 & 20.2 & 20.2 & 21.8 \\
& TEMPURA~\cite{nag2023unbiased} & 42.9 & 46.3 & 46.3 & 34.0 & 35.2 & 35.2 & 22.6 & 22.6 & 23.7 \\
& FloCoDe~\cite{khandelwal2024flocode} & 44.8 & 49.2 & 49.3 & 37.4 & 39.2 & 39.4 & 24.2 & 24.2 & 27.9 \\
& \textbf{VISA (Ours)} & \cellcolor{lightblue}\textbf{46.9} & \cellcolor{lightblue}\textbf{52.0} & \cellcolor{lightblue}\textbf{52.0} & \cellcolor{lightblue}\textbf{40.8} & \cellcolor{lightblue}\textbf{42.5} & \cellcolor{lightblue}\textbf{42.6} & \cellcolor{lightblue}\textbf{27.3} & \cellcolor{lightblue}\textbf{27.3} & \cellcolor{lightblue}\textbf{30.7} \\
\midrule

\multirow{2}{*}{\rotatebox{90}{Semi}} 
& TEMPURA~\cite{nag2023unbiased} & 40.7 & 44.5 & 44.6 & 36.9 & 39.5 & 39.5 & 21.8 & 21.8 & 22.5 \\
& \textbf{VISA (Ours)} & \cellcolor{lightblue}\textbf{51.3} & \cellcolor{lightblue}\textbf{56.3} & \cellcolor{lightblue}\textbf{56.4} & \cellcolor{lightblue}\textbf{47.8} & \cellcolor{lightblue}\textbf{52.6} & \cellcolor{lightblue}\textbf{52.6} & \cellcolor{lightblue}\textbf{31.7} & \cellcolor{lightblue}\textbf{31.7} & \cellcolor{lightblue}\textbf{33.2} \\
\midrule

\multirow{6}{*}{\rotatebox{90}{No}} 
& RelDN~\cite{zhang2019graphical} & 31.2 & 63.1 & 75.5 & 18.6 & 36.9 & 42.6 & 7.5 & 18.8 & 33.7 \\
& TRACE~\cite{teng2021target} & 50.9 & 73.6 & 82.7 & 31.9 & 42.7 & 46.3 & 22.8 & 31.3 & 41.8 \\
& STTran~\cite{cong2021spatial} & 51.4 & 67.7 & 82.7 & 40.7 & 50.1 & 58.8 & 20.9 & 29.7 & 39.2 \\ 
& TEMPURA~\cite{nag2023unbiased} & 61.5 & 85.1 & 98.0 & 48.3 & 61.1 & 66.4 & 24.7 & 33.9 & 43.7 \\
& FloCoDe~\cite{khandelwal2024flocode} & 63.2 & 86.9 & 98.6 & 49.7 & 63.8 & 69.2 & 28.6 & 35.4 & 47.2 \\
& \textbf{VISA (Ours)} & \cellcolor{lightblue}\textbf{65.8} & \cellcolor{lightblue}\textbf{89.1} & \cellcolor{lightblue}\textbf{99.8} & \cellcolor{lightblue}\textbf{52.0} & \cellcolor{lightblue}\textbf{66.3} & \cellcolor{lightblue}\textbf{71.4} & \cellcolor{lightblue}\textbf{30.7} & \cellcolor{lightblue}\textbf{36.7} & \cellcolor{lightblue}\textbf{50.4} \\
\bottomrule
\end{tabular}
\label{tab:sota}
\end{table*}

\begin{figure}[t]
  \centering
  \includegraphics[width=0.48\textwidth]{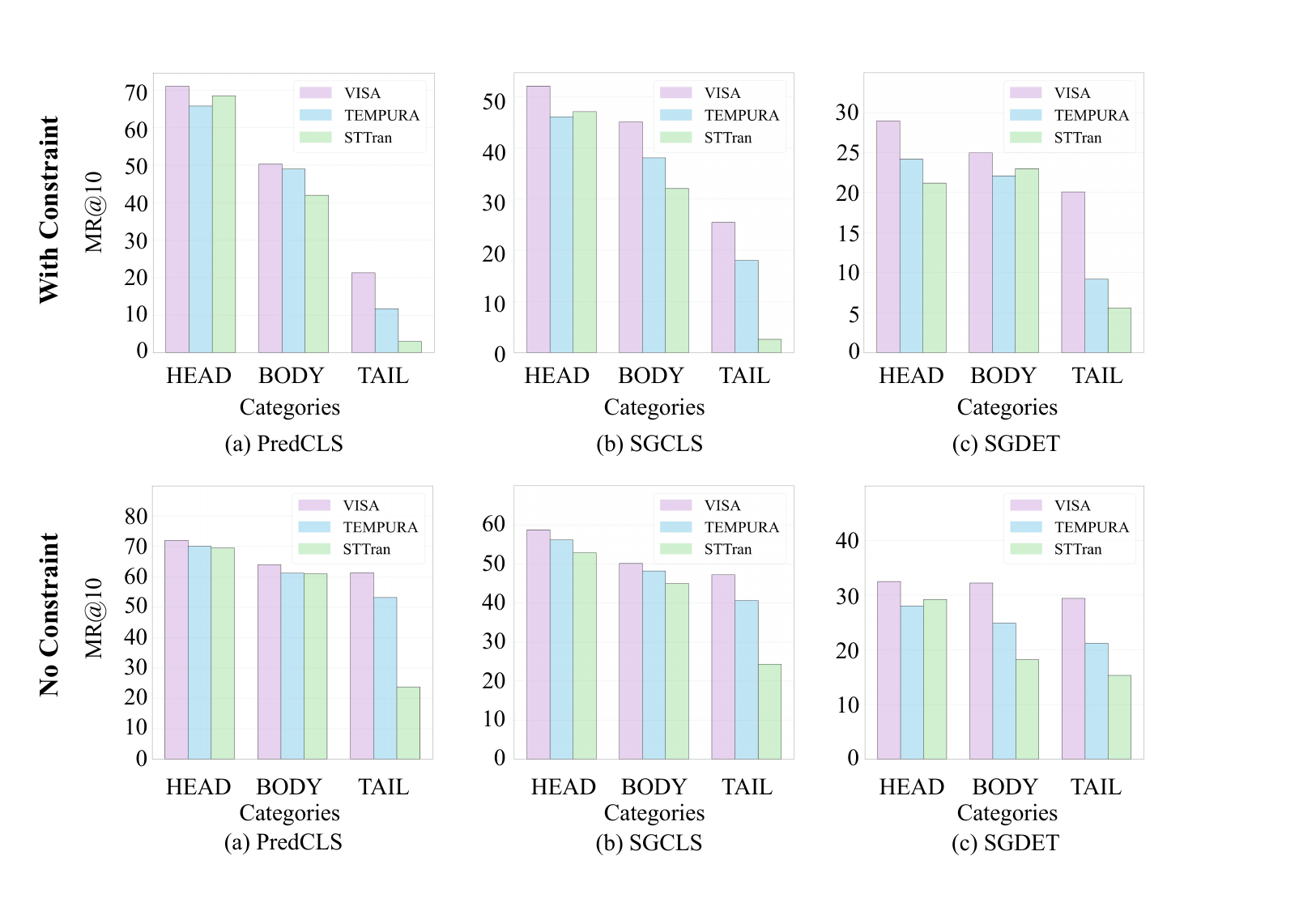}
  \caption{{\bf Quantitative results of our proposed method.} Compare our proposed method with open-sourced methods, TEMPURA~\cite{nag2023unbiased} and STTran~\cite{cong2021spatial} on the mR@10 metric under With Constraint and No Constraint. }
  \label{fig:mR@10}
  \vspace{-1em}
\end{figure}

\subsection{Comparison to existing methods}
\label{sec:SOTA}

We began with a qualitative analysis, shown in Fig.~\ref{fig:visualization}, illustrating results under the Semi Constraint condition, which closely resembles real-world scenarios~\cite{cong2021spatial} where multiple predicates typically describe scenes. False positive relationships and object classifications were highlighted in purple blocks. The attention intensity demonstrated that our framework effectively enhanced object representations (\eg, \textit{pillow}) through visual debiasing across frames from $t$ to $t+1$, compared with the TEMPURA~\cite{nag2023unbiased}. Additionally, our approach inferred more accurate semantic predicates (\eg, \textit{sit on}) within a single frame via contextual information. These visualizations confirm the correctness of partitioning scene graph biases into visual and semantic biases and demonstrate that our framework, VISA, significantly mitigates both in VidSGG.

Fig.~\ref{fig:visual} and Fig.~\ref{fig:matrix} further validated our theoretical analysis. Fig.~\ref{fig:visual} illustrated the variance of feature representations across training epochs, with shaded regions indicating the standard deviation. TEMPURA exhibited higher variance, suggesting less stable feature representations, whereas VISA demonstrated significantly reduced variance and tighter standard deviation regions. This reduction confirmed the effectiveness of the MGSM module in stabilizing feature representations. Additionally, Fig.~\ref{fig:matrix} presented the confusion matrix comparing relationship predictions between TEMPURA and VISA. The reduced influence from prior dataset biases in VISA validated our approach to mitigate the semantics bias in relationship predictions using KL divergence principles.

Our approach achieved state-of-the-art performance across all mR@K metrics, as shown in Table~\ref{tab:sota}. VISA demonstrated improvements ranging from a minimum of \textbf{+1.2\%} in mR@50 for the \textit{PREDCLS} task under the No Constraint scenario to a maximum of \textbf{+13.1\%} in both mR@20 and mR@50 for the \textit{SGCLS} task under the Semi Constraint scenario. These results validate the correctness of partitioning scene graph biases into visual and semantic biases and demonstrate the effectiveness of VISA in mitigating them, thereby advancing unbiased VidSGG.

\begin{table*}[t!]
\centering
\caption{Ablation study of MGSM and IRG under With Constraint, Semi Constraint, No Constraint.}
\vspace{-0.5em}
\scriptsize
\setlength{\tabcolsep}{0.6pt} 
\resizebox{1\linewidth}{!}{
\begin{tabular*}{\textwidth}{@{\extracolsep{\fill}} c cc cc cc cc cc cc cc cc cc cc}
\toprule
  &  \multicolumn{6}{c}{With Constraint} & \multicolumn{6}{c}{Semi Constraint} & \multicolumn{6}{c}{No Constraint}\\
  \cmidrule(lr){2-7} \cmidrule(lr){8-13} \cmidrule(lr){14-19}
    &
  \multicolumn{2}{c}{PREDCLS} & \multicolumn{2}{c}{SGCLS} & \multicolumn{2}{c}{SGDET} &
  \multicolumn{2}{c}{PREDCLS} & \multicolumn{2}{c}{SGCLS} & \multicolumn{2}{c}{SGDET} &
  \multicolumn{2}{c}{PREDCLS} & \multicolumn{2}{c}{SGCLS} & \multicolumn{2}{c}{SGDET} \\ 
  \cmidrule(lr){2-3} \cmidrule(lr){4-5} \cmidrule(lr){6-7} 
  \cmidrule(lr){8-9} \cmidrule(lr){10-11} \cmidrule(lr){12-13}
  \cmidrule(lr){14-15} \cmidrule(lr){16-17} \cmidrule(lr){18-19} 
  &  mR@10  & mR@20  & mR@10  & mR@20   & mR@10 & mR@20 &  mR@10  & mR@20  & mR@10  & mR@20   & mR@10 & mR@20 &  mR@10  & mR@20  & mR@10  & mR@20   & mR@10 & mR@20
\\
\midrule
\rowcolor{lightblue}
\multicolumn{1}{l}{\textbf{VISA}}  &   \textbf{46.9} &\textbf{52.0} &\textbf{40.8} &\textbf{42.5} & \textbf{27.3} &\textbf{27.3} &\textbf{51.3} &\textbf{56.3} & \textbf{47.8} &  \textbf{52.6} &\textbf{31.7} &\textbf{31.7} & \textbf{65.8} &\textbf{89.1} &\textbf{52.0} &\textbf{66.3}  & \textbf{30.7}& \textbf{36.7} \\

\multicolumn{1}{l}{w/o MGSM} & 46.9  & 52.0  & 35.7 & 38.2 & 18.8  & 24.6 & 51.3  & 56.3 & 45.6  & 49.3 & 24.1  & 27.8 & 65.8  & 89.1 & 49.0  & 62.0   & 27.9  & 34.7 \\
\multicolumn{1}{l}{w/o MGSM \& IRG}  & 42.9   & 46.3  & 34.0 & 35.2   & 18.5 & 22.6 & 40.7   & 44.5   & 36.9 & 39.5   & 18.5 & 21.8 &61.5 & 85.1 &48.3 & 61.1 & 24.7& 33.9 \\ 
\bottomrule
\end{tabular*}
}
\label{tab:abla}
\vspace{-0.8em}
\end{table*}

\begin{table*}[t!]
\centering
\caption{Influence of increasing iteration count \textit{N}.}
\vspace{-0.5em}
\scriptsize
\setlength{\tabcolsep}{0.6pt} 
\resizebox{1\linewidth}{!}{
\begin{tabular*}{\textwidth}{@{\extracolsep{\fill}} c cc cc cc cc cc cc cc cc cc cc}
\toprule
  &  \multicolumn{6}{c}{With Constraint} & \multicolumn{6}{c}{Semi Constraint} & \multicolumn{6}{c}{No Constraint}\\
  \cmidrule(lr){2-7} \cmidrule(lr){8-13} \cmidrule(lr){14-19}
    &
  \multicolumn{2}{c}{PREDCLS} & \multicolumn{2}{c}{SGCLS} & \multicolumn{2}{c}{SGDET} &
  \multicolumn{2}{c}{PREDCLS} & \multicolumn{2}{c}{SGCLS} & \multicolumn{2}{c}{SGDET} &
  \multicolumn{2}{c}{PREDCLS} & \multicolumn{2}{c}{SGCLS} & \multicolumn{2}{c}{SGDET} \\ 
  \cmidrule(lr){2-3} \cmidrule(lr){4-5} \cmidrule(lr){6-7} 
  \cmidrule(lr){8-9} \cmidrule(lr){10-11} \cmidrule(lr){12-13}
  \cmidrule(lr){14-15} \cmidrule(lr){16-17} \cmidrule(lr){18-19} 
  &  mR@10  & mR@20  & mR@10  & mR@20   & mR@10 & mR@20 &  mR@10  & mR@20  & mR@10  & mR@20   & mR@10 & mR@20 &  mR@10  & mR@20  & mR@10  & mR@20   & mR@10 & mR@20
\\
\midrule
\multicolumn{1}{l}{VISA}  &   46.9 &52.0&40.8 &42.5 & 27.3 &27.3 &51.3 &56.3 & 47.8 &  52.6 &31.7 &31.7 & 65.8 &89.1 &52.0 &66.3  & 30.7& 36.7 \\

\multicolumn{1}{l}{VISA$_{N=2}$} & 47.9 & 53.0 & 41.7 & 43.4 & 25.5 & 28.7 & 52.5 & 56.9 & 48.2 & 53.6 & 27.5 & 31.9 & 66.0 & 90.8 & 53.1 & 67.0 & 31.8 & 37.2 \\
\rowcolor{lightblue}
\multicolumn{1}{l}{VISA$_{N=3}$} &\textbf{48.1}& \textbf{53.3}& \textbf{42.2}& \textbf{44.0} &\textbf{25.8}&\textbf{28.9} &\textbf{52.8}& \textbf{57.1}& \textbf{48.5}& \textbf{53.9} &\textbf{27.8}&\textbf{32.1} &\textbf{66.2}& \textbf{91.2}& \textbf{53.5}& \textbf{67.2} &\textbf{32.0}&\textbf{37.5}&\\
\bottomrule
\end{tabular*}
}
\label{tab:iters}
\vspace{-0.5em}
\end{table*}

\begin{table*}[t!]
\centering
\caption{Different values of $\lambda$ on SGCLS and SGDET, we summarize the results in mean-Recall@K, with best in \textbf{bold}.}
\vspace{-0.5em}
\LARGE
\setlength{\tabcolsep}{5pt} 
\resizebox{1\linewidth}{!}{
\begin{tabular}{ccccccccccccccccccc} 
\toprule
& \multicolumn{6}{c}{With Constraint} & \multicolumn{6}{c}{Semi Constraint} & \multicolumn{6}{c}{No Constraint}\\
\cmidrule(lr){2-7} \cmidrule(lr){8-13} \cmidrule(lr){14-19}
$\lambda$ & 
\multicolumn{3}{c}{SGCLS} & \multicolumn{3}{c}{SGDET} & 
\multicolumn{3}{c}{SGCLS} & \multicolumn{3}{c}{SGDET} & 
\multicolumn{3}{c}{SGCLS} & \multicolumn{3}{c}{SGDET} \\ 
\cmidrule(lr){2-4} \cmidrule(lr){5-7}  \cmidrule(lr){8-10} \cmidrule(lr){11-13} 
\cmidrule(lr){14-16} \cmidrule(lr){17-19}
&mR@10 &mR@20 &mR@50 &mR@10 &mR@20 &mR@50 &mR@10  &mR@20 &mR@50 &mR@10 &mR@20 &mR@50 
&mR@10  &mR@20 &mR@50 &mR@10 &mR@20 &mR@50
\\
\midrule
0.005 &  34.2& 36.2& 36.2 & 17.9 & 21.5& 22.5  & 40.1 &	42.6 & 42.7 &21.5 & 26.6 & 28.1 & 46.4& 57.9& 57.9& 24.8& 32.5& 42.5 \\

0.01 &  36.0& 37.4& 37.4 & 18.6 & 22.2& 23.2  & 40.7 &	44.6 & 44.7 &23.7 & 28.6 & 30.0 & 47.2& 59.8& 65.2& 25.4& 34.0& 43.0 \\

0.02& 39.2 & 41.2 & 41.8 & 18.4 & 22.1& 23.0& 43.1& 47.0& 47.1   & 22.5 &  27.1&  28.1&48.3 &62.4 & 67.5& 24.5 &32.8 &41.7 \\
  
0.04& \cellcolor{lightblue}\textbf{40.8} & \cellcolor{lightblue}\textbf{42.5} & \cellcolor{lightblue}\textbf{42.6} & 19.8 & 23.3 & 24.3& \cellcolor{lightblue}\textbf{47.8}& \cellcolor{lightblue}\textbf{52.6} & \cellcolor{lightblue}\textbf{52.6}   & 24.5&  29.5& 31.1&\cellcolor{lightblue}\textbf{52.0} &\cellcolor{lightblue}\textbf{66.3} & \cellcolor{lightblue}\textbf{71.4}& 26.7 & 33.6 &41.7 \\
  
0.06& 35.1 & 36.4 & 36.4  & \cellcolor{lightblue}\textbf{27.3} & \cellcolor{lightblue}\textbf{27.3} & \cellcolor{lightblue}\textbf{30.7}& 41.8& 45.8 & 45.9   & \cellcolor{lightblue}\textbf{31.7} &  \cellcolor{lightblue}\textbf{31.7}&  \cellcolor{lightblue}\textbf{33.2}& 46.9 &59.3& 64.7&\cellcolor{lightblue}\textbf{30.7} & \cellcolor{lightblue}\textbf{36.7} &\cellcolor{lightblue}\textbf{50.4} \\
  
0.08& 34.3 & 35.6 & 35.6  & 18.3 & 21.6 & 22.5& 40.2& 43.7 & 43.7   & 23.3 &  27.8&  28.9&47.1 &59.6 & 65.7& 24.7 &32.7 &40.6 \\

0.1& 33.2 & 34.4 & 34.4  & 17.5 & 20.3 & 22.1& 39.1& 42.5 & 42.5   & 22.4 &  25.4&  27.2&46.2 &59.6 & 64.3& 22.1 &31.4 &39.2 \\
\bottomrule
\end{tabular}
}
\label{tab:k_no}
\vspace{-0.5em}
\end{table*}

\begin{table}[t]
\vspace{0.5em}
\caption{Performance evaluation on the PVSG~\cite{yang2023panoptic} Dataset}
\vspace{-0.5em}
\scriptsize
\resizebox{0.48\textwidth}{!}{ 
\begin{tabular}{ccccc}
\hline
\multicolumn{2}{c}{Method} & \multicolumn{3}{c}{PVSG Metrics} \\ \noalign{\vskip -0.3em} 
\cmidrule(lr){1-2} \cmidrule(lr){3-5}
Stage-1 & Stage-2 & mR@20 & mR@50 & mR@100 \\
\hline
VPS~\cite{cheng2022masked, li2022video} & Transformer & 0.36 & 0.39 & 0.40 \\
IPS+T~\cite{cheng2022masked, wang2021different} & Transformer & 1.75 & 1.86 & 2.03 \\
\hline
\rowcolor{lightblue}
\textbf{MGSM} & \textbf{IRG} & \textbf{8.82} & \textbf{8.94} & \textbf{10.51} \\
\hline
\end{tabular}
}
\vspace{0em}
\label{tab:pvsg}
\end{table}

\begin{table}[t]
    \caption{Performance evaluation on the PSG4D-GTA and PSG4D-HOI~\cite{yang2023psg4d} 4DPVSG Datasets.}
    \vspace{-0.5em}
    \resizebox{0.48\textwidth}{!}{ 
    \begin{tabular}{ccccccc}
        \toprule
        \multirow{2}{*}{\LARGE Method} & \multicolumn{3}{c}{\LARGE PSG4D-GTA} & \multicolumn{3}{c}{\LARGE PSG4D-HOI} \\
        \cmidrule(lr){2-4} \cmidrule(lr){5-7}
        & \LARGE mR@20 & \LARGE mR@50 & \LARGE mR@100 & \LARGE mR@20 & \LARGE mR@50 & \LARGE mR@100 \\
        \midrule
        \LARGE PSG4DFormer (P)~\cite{yang2023psg4d} & \LARGE 2.10 & \LARGE 2.93 & \LARGE 3.13 & \LARGE 3.10 & \LARGE 3.95 & \LARGE 4.17 \\
        \LARGE PSG4DFormer (R)~\cite{yang2023psg4d} & \LARGE 3.31 & \LARGE 3.85 & \LARGE 4.02 & \LARGE 3.65 & \LARGE 4.16 & \LARGE 4.97 \\
        \midrule
        \rowcolor{lightblue}
        \textbf{\LARGE MGSM + IRG} & \textbf{\LARGE 10.83} & \textbf{\LARGE 11.51} & \textbf{\LARGE 11.98} & \textbf{\LARGE 10.95} & \textbf{\LARGE 12.32} & \textbf{\LARGE 12.87} \\
        \bottomrule
    \end{tabular}
    }
    \vspace{-1em}
    \label{tab:4dpvsg}
\end{table}

Furthermore, as shown in Fig.~\ref{fig:mR@10}, our VISA model surpassed open-source unbiased VidSGG methods in mR@10 across the three predicate class distributions (HEAD, BODY, TAIL) on AG under both With Constraint and No Constraint settings. Here, \textit{HEAD} represented classes with \( \geq 100,000 \) training samples, \textit{BODY} with fewer than 100,000, and \textit{TAIL} with fewer than 8,000. VISA consistently outperformed all competitors across mR@10 metrics, achieving a notable \textbf{+11.0\%} improvement for the tail class in the SGDET task under No Constraint. This strong performance on tail classes highlighted the effectiveness of VISA's dual debiasing strategy.

\subsection{Ablation Studies}
\label{sec:Ablation}

\textbf{Ablation study on VISA modules.}~Table~\ref{tab:abla} demonstrated the impact of the visual and semantic debiasing modules. To assess their effectiveness, we first removed MGSM, resulting in a minimum decrease of \textbf{-2.2\%} in mR@10 for SGCLS under the Semi Constraint setting, underscoring the strong visual debiasing capability of MGSM. Similarly, excluding the IRG module yielded a minimum decrease of \textbf{-0.7\%} in mR@10 for SGCLS under the No Constraint setting, highlighting the effective use of IRG in mitigating semantics bias. These results confirm the correctness of partitioning scene graph biases into visual and semantic biases, and demonstrate that our framework, VISA, significantly mitigates both in VidSGG. Notably, MGSM was not applied to the PREDCLS task, as this task relied on ground-truth visual input, making the presence or absence of MGSM neutral to the results.

\noindent\textbf{Influence of increasing iteration count \textit{N}.}~We investigated how varying the iteration number \textit{N} affects our framework's performance. Due to the limited computing resources, we configured the iteration count \textit{N} to 2 and 3 in this experiment. The results of this setup, denoted as VISA$_{N=2}$, and VISA$_{N=3}$ were detailed in Table~\ref{tab:iters}. Upon examining the effect of iteration number $N$ on the performance of the unbiased VidSGG in terms of mR@K, it is evidenced that higher $N$ yields improved results. Notably, the performance gain from $N$ = 2 to $N$ = 3 is smaller than from $N$ = 1 to $N$ = 2. Additionally, setting $N$ = 2 doubled the training time compared to $N$ = 1. We provided additional details in the supplementary materials.

\noindent\textbf{The value of the $\lambda$.}~Table~\ref{tab:k_no} showed results for different values of $\lambda$. VISA achieved optimal performance for SGCLS at $\lambda = 0.04$, while the best results for SGDET occur at $\lambda = 0.06$. The differences in optimal $\lambda$ values reflect the distinct characteristics of these tasks. Specifically, \textit{SGCLS} benefits from ground-truth data, enabling effective updates to the memory representations with a smaller $\lambda$. In contrast, SGDET, which builds scene graphs from scratch, performs better with a slightly higher $\lambda$. Notably, since the MSGM module was not applied to the PREDCLS task, the parameter $\lambda$ is also not involved in PREDCLS.

\noindent\textbf{Exploration on PVSG and 4DPVSG Datasets.}~We further investigated the Panoptic Video Scene Graph Generation (PVSG) and 4D Panoptic Scene Graph Generation (4DPVSG) tasks, which are similar to VidSGG~\cite{yang2023panoptic,yang2023psg4d}. The key difference between PVSG and VidSGG was the grounding of each node, PVSG uses pixel-level segmentation masks, while VidSGG uses bounding boxes. Using the IPS+T+Transformer codebase, we employed visual and semantic debiasing strategies as previously described. We completed the MGSM framework with a Memory-Enhanced Temporal Integrator and linked frames with UniTrack~\cite{wang2021different} to generate tracked video cubes. During relation classification, we adapted IRG for iterative semantic debiasing. Similarly, the 4DPVSG task extended PVSG to a dynamic 4D context. PSG4DFormer (P) and PSG4DFormer (R) denote the model variants of input point cloud and RGB-D sequences, respectively. As shown in Table~\ref{tab:pvsg}, our approach significantly outperformed the baseline, with a notable \textbf{minimum} gain of \textbf{+7.52\%} in mR@20 for PVSG and \textbf{+7.30\%} for 4DPVSG in mR@20 for PSG4D-HOI datasets. These results demonstrated the general applicability of our dual debiasing approach across different VidSGG tasks, potentially benefiting other generative tasks and applications in embodied intelligence.
\section{Conclusion}
\vspace{-0.5em}
\label{sec:conclusion}
In this work, we present VISA, an approach for unbiased video scene graph generation that addresses both visual and semantic biases inherent in scene graphs. VISA reduces visual bias by enriching current frame representations with historical context to decrease visual feature variance, and mitigates semantic bias by increasing the KL divergence between model predictions and biased priors. Our framework integrates hierarchical semantic contexts with visual features, enabling more unbiased scene graph generation. Extensive experimental and theoretical analyses demonstrate the effectiveness of VISA. Future work will explore extending our visual-semantic debiasing approach to other multimodal domains such as visual question answering and video captioning, as well as applications in embodied intelligence, including robotics perception and interactive agents.
\section*{Acknowledgments}
We gratefully acknowledge the three anonymous reviewers for their constructive feedback on our CVPR 2025 submission. This work was supported by personal funding from the corresponding author.

\clearpage
\setcounter{page}{1}
\maketitlesupplementary

\appendix
\renewcommand\thepage{S\arabic{page}} 
\setcounter{section}{0} 
\renewcommand\thesection{\Alph{section}} 

\section{Additional Details on MGSM}
\label{sec:Additional_MGSM}

In this section, we derive the variance of the memory representation $\mathbf{M}_i^t$ and establish the upper and lower bounds for the update parameter $\lambda$. In our proposed Memory Guided Sequence Modeling (MGSM) module, the memory representation \(\mathbf{M}_i^t\) is updated using the following equation:
\vspace{-0.2em}
\begin{equation}
    \mathbf{M}_i^{t+1} = (1 - \lambda) \mathbf{M}_i^t + \lambda \mathbf{v}_i^t,
    \label{eq:memory_update}
\end{equation}

\noindent where \(\lambda\) is the update rate, and \(\mathbf{v}_i^t\) represents the feature vector at time step \(t\) for object \(i\). The feature vector \(\mathbf{v}_i^t\) is modeled as:
\vspace{-0.5em}
\begin{equation}
    \mathbf{v}_i^t = \mathbf{v}_i + \boldsymbol{\epsilon}_i^t,
    \label{eq:feature_model}
\end{equation}

\noindent with \(\boldsymbol{\epsilon}_i^t\) being zero-mean Gaussian noise with covariance \(\boldsymbol{\Sigma}\), i.e.,
\vspace{-0.5em}
\begin{equation}
    \mathbb{E}[\boldsymbol{\epsilon}_i^t] = \mathbf{0}, \quad \operatorname{Cov}[\boldsymbol{\epsilon}_i^t] = \boldsymbol{\Sigma}.
    \label{eq:noise_properties}
\end{equation}

We assume that the noise terms are independent across different time steps and objects. To derive the variance of the memory representation \(\mathbf{M}_i^t\), we proceed as follows.
\vspace{-1em}
\paragraph{Expectation of Memory Representation}

Taking the expectation of both sides of the update equation \eqref{eq:memory_update}:
\vspace{-0.2em}
\begin{equation}
    \mathbb{E}[\mathbf{M}_i^{t+1}] = (1 - \lambda) \mathbb{E}[\mathbf{M}_i^t] + \lambda \mathbb{E}[\mathbf{v}_i^t].
    \label{eq:expectation_update}
\end{equation}

\noindent Since \(\mathbb{E}[\boldsymbol{\epsilon}_i^t] = \mathbf{0}\) from equation \eqref{eq:noise_properties}, we have:
\vspace{-0.5em}
\begin{equation}
    \mathbb{E}[\mathbf{v}_i^t] = \mathbf{v}_i.
    \label{eq:expectation_feature}
\end{equation}

Assuming steady-state where \(\mathbb{E}[\mathbf{M}_i^{t+1}] = \mathbb{E}[\mathbf{M}_i^t] = \mathbf{M}\), equation \eqref{eq:expectation_update} simplifies to:

\begin{equation}
    \mathbf{M} = (1 - \lambda) \mathbf{M} + \lambda \mathbf{v}_i \quad \Rightarrow \quad \mathbf{M} = \mathbf{v}_i.
    \label{eq:steady_state_expectation}
\end{equation}
\vspace{-2.5em}
\paragraph{Variance of Memory Representation}

Next, we compute the variance \(\operatorname{Var}[\mathbf{M}_i^t]\). Taking the variance of both sides of the update equation \eqref{eq:memory_update}:
\vspace{-0.5em}
\begin{equation}
    \operatorname{Var}[\mathbf{M}_i^{t+1}] = \operatorname{Var}[(1 - \lambda) \mathbf{M}_i^t + \lambda \mathbf{v}_i^t].
    \label{eq:variance_update}
\end{equation}

\noindent Since \(\mathbf{M}_i^t\) and \(\mathbf{v}_i^t\) are independent, the variance propagates as:
\vspace{-0.5em}
\begin{equation}
    \operatorname{Var}[\mathbf{M}_i^{t+1}] = (1 - \lambda)^2 \operatorname{Var}[\mathbf{M}_i^t] + \lambda^2 \operatorname{Var}[\mathbf{v}_i^t].
    \label{eq:variance_propagation}
\end{equation}

\noindent Given that \(\operatorname{Var}[\mathbf{v}_i^t] = \boldsymbol{\Sigma}\) from equation \eqref{eq:noise_properties}, we substitute into equation \eqref{eq:variance_propagation}:
\vspace{-0.5em}
\begin{equation}
    \operatorname{Var}[\mathbf{M}_i^{t+1}] = (1 - \lambda)^2 \operatorname{Var}[\mathbf{M}_i^t] + \lambda^2 \boldsymbol{\Sigma}.
    \label{eq:variance_substitution}
\end{equation}

\noindent Assuming steady-state where \(\operatorname{Var}[\mathbf{M}_i^{t+1}] = \operatorname{Var}[\mathbf{M}_i^t] = \mathbf{V}\), we get:

\begin{equation}
    \mathbf{V} = (1 - \lambda)^2 \mathbf{V} + \lambda^2 \boldsymbol{\Sigma}.
    \label{eq:steady_state_variance}
\end{equation}

\noindent Solving for \(\mathbf{V}\):
\vspace{-0.5em}
\begin{align}
    \mathbf{V} \left[1 - (1 - \lambda)^2\right] &= \lambda^2 \boldsymbol{\Sigma}, \\
    1 - (1 - \lambda)^2 &= 2\lambda - \lambda^2, \\
    \mathbf{V} (2\lambda - \lambda^2) &= \lambda^2 \boldsymbol{\Sigma}, \\
    \mathbf{V} &= \frac{\lambda^2 \boldsymbol{\Sigma}}{2\lambda - \lambda^2} = \frac{\lambda \boldsymbol{\Sigma}}{2 - \lambda}.
    \label{eq:variance_solution}
\end{align}

\noindent For small \(\lambda\), the expression simplifies to:

\begin{equation}
    \operatorname{Var}[\mathbf{M}_i^t] = \frac{\lambda \boldsymbol{\Sigma}}{2 - \lambda} \approx \frac{\lambda \boldsymbol{\Sigma}}{2}.
    \label{eq:variance_approx}
\end{equation}

While minimizing the variance of the memory representation is desirable for enhancing stability, the update parameter \(\lambda\) must be carefully selected to balance variance reduction and the model's ability to adapt to new information. Specifically, \(\lambda\) cannot be too small, as excessively small values slow the adaptation to new information and may introduce significant bias.
\vspace{-1.5em}
\paragraph{Bias-Variance Trade-off}

The total error in the memory representation can be decomposed into bias and variance:

\begin{equation}
    \text{Total Error} = \text{Bias}^2 + \text{Variance}.
    \label{eq:total_error}
\end{equation}

\noindent From the variance derivation in equation \eqref{eq:variance_approx}, we have:

\begin{equation}
    \operatorname{Var}[\mathbf{M}_i^t] \approx \frac{\lambda \boldsymbol{\Sigma}}{2}.
    \label{eq:variance_final}
\end{equation}

\noindent Next, we analyze the bias introduced by the update mechanism. Assume that the feature vector evolves over time as:

\begin{equation}
    \mathbf{v}_i^{t+1} = \mathbf{v}_i^t + \boldsymbol{\delta},
    \label{eq:feature_evolution}
\end{equation}

\noindent where \(\boldsymbol{\delta}\) is a constant change vector representing the feature change rate. Substituting equation \eqref{eq:feature_evolution} into the memory update equation \eqref{eq:memory_update}, we get:

\begin{equation}
    \mathbf{M}_i^{t+1} = (1 - \lambda) \mathbf{M}_i^t + \lambda (\mathbf{v}_i^t + \boldsymbol{\delta}).
    \label{eq:memory_update_evolution}
\end{equation}

Assuming no noise for bias analysis (\(\boldsymbol{\epsilon}_i^t = \mathbf{0}\)) and iteratively applying the update equation starting from \(t = 0\), we can derive the bias over time.

\paragraph{Bias Derivation}

At \(t = 0\), the initial memory is set to the initial feature:
\vspace{-0.5em}
\begin{equation}
    \mathbf{M}_i^0 = \mathbf{v}_i^0 = \mathbf{v}_i.
    \label{eq:initial_memory}
\end{equation}

\noindent For \(t \geq 0\), the update equation without noise becomes:
\begin{equation}
    \mathbf{M}_i^{t+1} = (1 - \lambda) \mathbf{M}_i^t + \lambda (\mathbf{v}_i^t + \boldsymbol{\delta}).
    \label{eq:memory_update_no_noise}
\end{equation}

\noindent Substituting the feature evolution from equation 
\vspace{-0.5em}
\eqref{eq:feature_evolution}:

\begin{equation}
    \mathbf{v}_i^t = \mathbf{v}_i^{t-1} + \boldsymbol{\delta} = \mathbf{v}_i + t \boldsymbol{\delta}.
    \label{eq:feature_evolution_expanded}
\end{equation}

\noindent Thus, the update equation becomes:
\begin{align}
    \mathbf{M}_i^{t+1} &= (1 - \lambda) \mathbf{M}_i^t + \lambda (\mathbf{v}_i + t \boldsymbol{\delta} + \boldsymbol{\delta}) \nonumber \\
    &= (1 - \lambda) \mathbf{M}_i^t + \lambda \mathbf{v}_i + \lambda (t + 1) \boldsymbol{\delta}.
    \label{eq:memory_update_expanded}
\end{align}

\noindent We can unroll this recurrence relation to find the general expression for \(\mathbf{M}_i^t\):
\begin{align}
    \mathbf{M}_i^t &= (1 - \lambda)^t \mathbf{M}_i^0 \nonumber \\
    &\quad + \lambda \sum_{k=0}^{t-1} (1 - \lambda)^k \mathbf{v}_i^{t - k - 1} \nonumber \\
    &\quad + \lambda \sum_{k=0}^{t-1} (1 - \lambda)^k \boldsymbol{\delta}.
    \label{eq:memory_unroll}
\end{align}

\noindent Since \(\mathbf{M}_i^0 = \mathbf{v}_i\), and \(\mathbf{v}_i^{t} = \mathbf{v}_i + t \boldsymbol{\delta}\), the expression simplifies over multiple iterations.
\vspace{-1em}
\paragraph{Steady-State Bias}

As \(t \rightarrow \infty\), the influence of the initial memory and transient terms diminishes, leading to a steady-state bias. From equation \eqref{eq:memory_update_evolution}, in steady-state, we have:
\vspace{-0.2em}
\begin{equation}
    \mathbf{M} = (1 - \lambda) \mathbf{M} + \lambda \mathbf{v}_i + \lambda \boldsymbol{\delta}.
    \label{eq:steady_state_bias}
\end{equation}

\noindent Solving for \(\mathbf{M}\):
\vspace{-0.5em}
\begin{equation}
    \lambda \boldsymbol{\delta} = \lambda (\mathbf{v}_i - \mathbf{M}) \quad \Rightarrow \quad \mathbf{M} = \mathbf{v}_i - \boldsymbol{\delta}.
    \label{eq:steady_state_bias_solution}
\end{equation}

\noindent Thus, the bias in the memory representation at steady-state is:
\vspace{-0.5em}
\begin{equation}
    \text{Bias} = \mathbf{M} - \mathbf{v}_i = -\boldsymbol{\delta}.
    \label{eq:bias}
\end{equation}

However, this simplistic analysis overlooks the dynamic nature of \(\mathbf{M}_i^t\). A more rigorous approach considers the cumulative effect of \(\lambda\) over time, leading to a residual bias that depends inversely on \(\lambda\).
\vspace{-1em}
\paragraph{Alternative Bias Derivation}

Assuming that at each time step, the feature vector increases by \(\boldsymbol{\delta}\), the memory update equation becomes:
\vspace{-0.2em}
\begin{align}
    \mathbf{M}_i^{t+1} &= (1 - \lambda) \mathbf{M}_i^t + \lambda (\mathbf{v}_i + t \boldsymbol{\delta} + \boldsymbol{\delta}) \nonumber \\
    &= (1 - \lambda) \mathbf{M}_i^t + \lambda \mathbf{v}_i + \lambda (t + 1) \boldsymbol{\delta}.
    \label{eq:memory_update_t_plus_1}
\end{align}

\noindent Unfolding this recursion, we find that the bias accumulates over time and converges to:

\begin{equation}
    \text{Bias}_{\infty} = \lim_{t \to \infty} (\mathbf{M}_i^t - \mathbf{v}_i^t) = -\frac{\boldsymbol{\delta}}{\lambda}.
    \label{eq:steady_state_bias_final}
\end{equation}

\noindent This shows that the steady-state bias is inversely proportional to \(\lambda\).
\vspace{-1em}
\paragraph{Lower Bound for \(\lambda\)}

From equation \eqref{eq:steady_state_bias_final}, we observe that:

\begin{equation}
    \text{Bias}_{\infty} = -\frac{\boldsymbol{\delta}}{\lambda}.
    \label{eq:bias_relation}
\end{equation}

As \(\lambda\) decreases, the magnitude of \(\text{Bias}_{\infty}\) increases. To ensure that the bias remains within acceptable limits, \(\lambda\) must be bounded below by a positive value. Specifically, to maintain \(\|\text{Bias}_{\infty}\| \leq \epsilon\), where \(\epsilon\) is the maximum tolerable bias, we derive:
\vspace{-0.2em}
\begin{equation}
    \left\| -\frac{\boldsymbol{\delta}}{\lambda} \right\| \leq \epsilon \quad \Rightarrow \quad \lambda \geq \frac{\|\boldsymbol{\delta}\|}{\epsilon}.
    \label{eq:lambda_lower_bound}
\end{equation}

\noindent Therefore, the lower bound for \(\lambda\) is:

\vspace{-0.2em}
\begin{equation}
    \lambda \geq \frac{\|\boldsymbol{\delta}\|}{\epsilon}.
    \label{eq:lambda_min}
\end{equation}

This implies that \(\lambda\) cannot be arbitrarily small, as doing so would result in an unbounded increase in bias, thereby compromising the accuracy and reliability of the memory representation.
\vspace{-1em}
\paragraph{Optimal \(\lambda\)}

To minimize the total error, which comprises both bias and variance, we balance the two components. From equations \eqref{eq:bias_relation} and \eqref{eq:variance_final}, the total error is:
\vspace{-0.5em}
\begin{equation}
    \text{Total Error} = \left\| \text{Bias}_{\infty} \right\|^2 + \operatorname{Var}[\mathbf{M}_i^t] = \frac{\|\boldsymbol{\delta}\|^2}{\lambda^2} + \frac{\lambda \boldsymbol{\Sigma}}{2}.
    \label{eq:total_error_full}
\end{equation}
\noindent To find the optimal \(\lambda\), we take the derivative of the total error with respect to \(\lambda\) and set it to zero:
\begin{equation}
    \frac{d}{d\lambda} \left( \frac{\|\boldsymbol{\delta}\|^2}{\lambda^2} + \frac{\lambda \boldsymbol{\Sigma}}{2} \right) = -2 \frac{\|\boldsymbol{\delta}\|^2}{\lambda^3} + \frac{\boldsymbol{\Sigma}}{2} = 0.
    \label{eq:derivative_total_error_full}
\end{equation}

\noindent Solving for \(\lambda\):
\vspace{-1em}
\begin{align}
    -2 \frac{\|\boldsymbol{\delta}\|^2}{\lambda^3} + \frac{\boldsymbol{\Sigma}}{2} &= 0, \\
    2 \frac{\|\boldsymbol{\delta}\|^2}{\lambda^3} &= \frac{\boldsymbol{\Sigma}}{2}, \\
    \lambda^3 &= \frac{4 \|\boldsymbol{\delta}\|^2}{\boldsymbol{\Sigma}}, \\
    \lambda &= \left( \frac{4 \|\boldsymbol{\delta}\|^2}{\boldsymbol{\Sigma}} \right)^{\frac{1}{3}}.
    \label{eq:optimal_lambda}
\end{align}

\noindent Thus, the optimal \(\lambda\) that minimizes the total error is:
\begin{equation}
    \lambda_{\text{opt}} = \left( \frac{4 \|\boldsymbol{\delta}\|^2}{\boldsymbol{\Sigma}} \right)^{\frac{1}{3}}.
    \label{eq:lambda_opt}
\end{equation}
\vspace{-1.5em}

\begin{table*}[h!]
\vspace{-0.5em}
\centering
\caption{Ablation study of MGSM and IRG under With Constraint.}
\vspace{-0.5em}
\small
\setlength{\tabcolsep}{2pt} 
\resizebox{1\linewidth}{!}{
\begin{tabular*}{\textwidth}{@{\extracolsep{\fill}} c ccc ccc ccc}
\toprule
  &  \multicolumn{9}{c}{With Constraint}\\
  \cmidrule(lr){2-10}
    &
  \multicolumn{3}{c}{PREDCLS} & \multicolumn{3}{c}{SGCLS} & \multicolumn{3}{c}{SGDET} \\ 
  \cmidrule(lr){2-4} \cmidrule(lr){5-7} \cmidrule(lr){8-10} 
  &  mR@10  & mR@20  & mR@50 & mR@10  & mR@20  & mR@50  & mR@10 & mR@20  & mR@50 
\\
\midrule
\multicolumn{1}{l}{\textbf{VISA}}  & \textbf{46.9} &\textbf{52.0} & \textbf{52.0}&\textbf{40.8} &\textbf{42.5} &\textbf{42.6}  & \textbf{27.3} &\textbf{27.3} & \textbf{30.7} \\
\hline

\multicolumn{1}{l}{w/o MGSM} & 46.9  & 52.0  & 52.0 & 35.7   & 38.2   & 38.2 & 18.8  & 24.6 & 26.5\\
\multicolumn{1}{l}{w/o MGSM}\& IRG  & 42.9   & 46.3   & 46.3   & 34.0 & 35.2  & 35.2   & 18.5 & 22.6 &23.1 \\ 

\bottomrule
\end{tabular*}
}
\label{tab:abla_with}
\end{table*}

\begin{table*}[ht]
\centering
\caption{Ablation study of MGSM and IRG under Semi Constraint.}
\vspace{-0.5em}
\small
\setlength{\tabcolsep}{2pt} 
\resizebox{1\linewidth}{!}{
\begin{tabular*}{\textwidth}{@{\extracolsep{\fill}} c ccc ccc ccc}
\toprule
  &  \multicolumn{9}{c}{Semi Constraint}\\
  \cmidrule(lr){2-10}
    &
  \multicolumn{3}{c}{PREDCLS} & \multicolumn{3}{c}{SGCLS} & \multicolumn{3}{c}{SGDET} \\ 
  \cmidrule(lr){2-4} \cmidrule(lr){5-7} \cmidrule(lr){8-10} 
  &  mR@10  & mR@20  & mR@50 & mR@10  & mR@20  & mR@50  & mR@10 & mR@20  & mR@50 
\\
\midrule
\multicolumn{1}{l}{\textbf{VISA}}  & \textbf{51.3} & \textbf{56.3}  & \textbf{56.4}  & \textbf{47.8}  & \textbf{52.6} & \textbf{52.6}  &  \textbf{31.7} & \textbf{31.7}& \textbf{33.2} \\
\hline
\multicolumn{1}{l}{w/o MGSM} & 51.3  & 56.3  & 56.4 & 44.6  & 48.4  & 48.6   & 23.1  & 27.5 & 28.5\\
\multicolumn{1}{l}{w/o MGSM}\& IRG  & 40.7   & 44.5  & 44.6  & 36.9 & 39.5  & 39.5   & 18.5 & 21.8 &22.5 \\ 

\bottomrule
\end{tabular*}
}
\label{tab:abla_semi}
\end{table*}

\begin{table*}[t!]
\centering
\caption{Ablation study of MGSM and IRG under No Constraint.}
\vspace{-0.5em}
\small
\setlength{\tabcolsep}{2pt} 
\resizebox{1\linewidth}{!}{
\begin{tabular*}{\textwidth}{@{\extracolsep{\fill}} c ccc ccc ccc}
\toprule
  &  \multicolumn{9}{c}{No Constraint}\\
  \cmidrule(lr){2-10}
    &
  \multicolumn{3}{c}{PREDCLS} & \multicolumn{3}{c}{SGCLS} & \multicolumn{3}{c}{SGDET} \\ 
  \cmidrule(lr){2-4} \cmidrule(lr){5-7} \cmidrule(lr){8-10} 
  &  mR@10  & mR@20  & mR@50 & mR@10  & mR@20  & mR@50  & mR@10 & mR@20  & mR@50 
\\
\midrule
\multicolumn{1}{l}{\textbf{VISA}}&  \textbf{65.8} &\textbf{89.1} & \textbf{99.8} &\textbf{52.0} &\textbf{66.3}  &\textbf{71.4}  & \textbf{30.7}& \textbf{36.7}& \textbf{50.4} \\
\hline
\multicolumn{1}{l}{w/o MGSM} & 65.8  & 89.1  & 99.8 & 49.0  & 62.0  & 67.2  & 27.9  & 34.7 & 47.2\\
\multicolumn{1}{l}{w/o MGSM}\& IRG  & 61.5   & 85.1  & 95.9  & 48.3 & 61.1  & 66.0   & 24.7 & 33.9 &45.9 \\ 

\bottomrule
\end{tabular*}
}
\label{tab:abla_no}
\end{table*}

\subsection{Empirical Validation}

In practice, the optimal \(\lambda\) is determined based on the specific values of \(\boldsymbol{\delta}\) and \(\boldsymbol{\Sigma}\) derived from the data. From the previous calculation, we get the approximate $\lambda$ value as following:
\begin{equation}
    \lambda \approx 0.04,
    \label{eq:empirical_lambda}
\end{equation}

\noindent it suggests that:
\vspace{-0.5em}
\begin{equation}
    \lambda = \left( \frac{4 \|\boldsymbol{\delta}\|^2}{\boldsymbol{\Sigma}} \right)^{\frac{1}{3}} \approx 0.04.
    \label{eq:lambda_empirical}
\end{equation}

This optimal value balances the trade-off between minimizing bias and controlling variance, ensuring robust feature estimation in the MGSM module.

Through the derivation, we establish that the variance of the memory representation decreases with a smaller \(\lambda\), enhancing stability, while the bias increases inversely with \(\lambda\), reducing adaptability. The optimal \(\lambda\) of approximately 0.04 in our experiments effectively balances this trade-off, providing both robust and adaptable feature representations essential for mitigating visual bias in video scene graph generation.

\section{Additional Metrics and Evaluation Setup}

\noindent We evaluate our approach using the standard metric for unbiased VidSGG, mean Recall@K (mR@K) with $K \in \{10, 20, 50\}$. TEMPURA~\cite{nag2023unbiased} serves as our baseline method. Following established protocols~\cite{ji2020action,cong2021spatial,nag2023unbiased}, we conduct three Scene Graph Generation (SGG) tasks:

\textbf{Predicate Classification} (\textit{PREDCLS}) which delivers object localization and classes, necessitating the model to discern predicate classes. 
\textbf{Scene Graph Classification} (\textit{SGCLS}) furnishes precise localization, expecting the model to identify both object and predicate classes. 
\textbf{Scene Graph Detection} (\textit{SGDET}) requires the model to initially detect bounding boxes before classifying objects and predicate classes.~Evaluation is conducted across three distinct settings: \textbf{With Constraint}, \textbf{Semi Constraint}, and \textbf{No Constraint}.~Under the With Constraint setting, the generated scene graphs are limited to at most one predicate per subject-object pair. The Semi Constraint setting allows for multiple predicates, yet only those surpassing a specified confidence threshold (=0.9) are considered. Scene graphs can contain multiple predicates between pairs without any restrictions under the No Constraint. It is important to emphasize that another standard metric R@K is susceptible to a bias favoring predominant predicate classes~\cite{tang2019learning,nag2023unbiased}, whereas mR@K is averaged across all predicate classes, thus providing an indicator of a model's performance on the unbiased VidSGG. Consequently, for the task of generating unbiased VidSGG, we will afford greater scrutiny to the mR@K metric, as it offers a more balanced assessment of model performance.

\begin{table*}[h!]
\vspace{-0.5em}
\centering
\caption{Ablation study of HSE under With Constraint.}
\vspace{-0.5em}
\small
\setlength{\tabcolsep}{2pt} 
\resizebox{1\linewidth}{!}{
\begin{tabular*}{\textwidth}{@{\extracolsep{\fill}} c ccc ccc ccc}
\toprule
  &  \multicolumn{9}{c}{With Constraint}\\
  \cmidrule(lr){2-10}
    &
  \multicolumn{3}{c}{PREDCLS} & \multicolumn{3}{c}{SGCLS} & \multicolumn{3}{c}{SGDET} \\ 
  \cmidrule(lr){2-4} \cmidrule(lr){5-7} \cmidrule(lr){8-10} 
  &  mR@10  & mR@20  & mR@50 & mR@10  & mR@20  & mR@50  & mR@10 & mR@20  & mR@50 
\\
\midrule
\multicolumn{1}{l}{\textbf{VISA}}  & \textbf{46.9} &\textbf{52.0} & \textbf{52.0}&\textbf{40.8} &\textbf{42.5} &\textbf{42.6}  & \textbf{27.3} &\textbf{27.3} & \textbf{30.7} \\
\hline

\multicolumn{1}{l}{w/o HSE} & 44.5  & 49.2  & 49.2 & 38.4 & 40.2   & 40.3 & 25.8  & 25.6 & 28.3\\

\bottomrule
\end{tabular*}
}
\label{tab:abla_HSE_with}
\end{table*}

\begin{table*}[ht]
\centering
\caption{Ablation study of HSE under Semi Constraint.}
\vspace{-0.5em}
\small
\setlength{\tabcolsep}{2pt} 
\resizebox{1\linewidth}{!}{
\begin{tabular*}{\textwidth}{@{\extracolsep{\fill}} c ccc ccc ccc}
\toprule
  &  \multicolumn{9}{c}{Semi Constraint}\\
  \cmidrule(lr){2-10}
    &
  \multicolumn{3}{c}{PREDCLS} & \multicolumn{3}{c}{SGCLS} & \multicolumn{3}{c}{SGDET} \\ 
  \cmidrule(lr){2-4} \cmidrule(lr){5-7} \cmidrule(lr){8-10} 
  &  mR@10  & mR@20  & mR@50 & mR@10  & mR@20  & mR@50  & mR@10 & mR@20  & mR@50 
\\
\midrule
\multicolumn{1}{l}{\textbf{VISA}}  & \textbf{51.3} & \textbf{56.3}  & \textbf{56.4}  & \textbf{47.8}  & \textbf{52.6} & \textbf{52.6}  &  \textbf{31.7} & \textbf{31.7}& \textbf{33.2} \\
\hline
\multicolumn{1}{l}{w/o HSE} & 49.1  & 54.2  & 54.6 & 45.4  & 50.4  & 50.4   & 29.1  & 29.1 & 31.5\\

\bottomrule
\end{tabular*}
}
\label{tab:abla_HSE_semi}
\end{table*}

\begin{table*}[t!]
\centering
\caption{Ablation study of HSE under No Constraint.}
\vspace{-0.5em}
\small
\setlength{\tabcolsep}{2pt} 
\resizebox{1\linewidth}{!}{
\begin{tabular*}{\textwidth}{@{\extracolsep{\fill}} c ccc ccc ccc}
\toprule
  &  \multicolumn{9}{c}{No Constraint}\\
  \cmidrule(lr){2-10}
    &
  \multicolumn{3}{c}{PREDCLS} & \multicolumn{3}{c}{SGCLS} & \multicolumn{3}{c}{SGDET} \\ 
  \cmidrule(lr){2-4} \cmidrule(lr){5-7} \cmidrule(lr){8-10} 
  &  mR@10  & mR@20  & mR@50 & mR@10  & mR@20  & mR@50  & mR@10 & mR@20  & mR@50 
\\
\midrule
\multicolumn{1}{l}{\textbf{VISA}}&  \textbf{65.8} &\textbf{89.1} & \textbf{99.8} &\textbf{52.0} &\textbf{66.3}  &\textbf{71.4}  & \textbf{30.7}& \textbf{36.7}& \textbf{50.4} \\
\hline
\multicolumn{1}{l}{w/o HSE} & 63.3  & 87.0  & 98.4 & 50.0  & 64.0  & 69.2  & 27.9  & 34.7 & 48.5\\

\bottomrule
\end{tabular*}
}
\label{tab:abla_HSE_no}
\end{table*}

\section{Additional Implementation Details}
\label{sec:implementation}
Following prior work~\cite{cong2021spatial,li2022dynamic,nag2023unbiased}, we adopted Faster R-CNN~\cite{ren2015faster} with ResNet-101~\cite{he2016deep} as the object detector, initially trained on the AG dataset. To ensure a fair comparison, we utilized the official implementations of these methods. For our MGSM module, we set the $\lambda$ parameter to $0.04$ for the \textit{SGCLS} task and $0.06$ for the \textit{SGDET} task. In the IRG module, we implemented a dual-procedure setup, enabling iterative relational inference with the number of iterations $N$ set to $1$. The framework was trained end to end for $15$ epochs using the AdamW optimizer~\cite{loshchilov2017decoupled} and a batch size of $1$. The initial learning rate was $10^{-5}$. We reduced the initial learning rate by 0.5 whenever the performance plateaus. All code ran on a single RTX 4090.

\section{Supplementary Experimental Results}
\subsection{Complete Ablation study on VISA modules}
Due to page constraints, the mR@50 results for the ablation study on VISA modules were omitted from the main text. Here, we present the complete ablation study in Tables~\ref{tab:abla_with},~\ref{tab:abla_semi}, and~\ref{tab:abla_no}, demonstrating the effects of the visual and semantic debiasing modules. 
Consistent with the ablation study in the main body, we first removed the MGSM module. Focusing on the mR@50 results, this removal led to a minimal decrease of \textbf{-3.2\%} in mR@50 for SGDET under the No Constraint setting, underscoring MGSM's strong visual debiasing capability. Similarly, excluding the IRG module resulted in a minimal decrease of \textbf{-1.3\%} in mR@50 for SGDET under the No Constraint setting, highlighting IRG's effectiveness in mitigating semantic bias. These findings validate our approach of partitioning scene graph biases into visual and semantic components, demonstrating that our VISA framework effectively mitigates both biases in VidSGG. Notably, MGSM was not applied to the PREDCLS task, as this task relies on ground-truth visual input, rendering the inclusion of MGSM neutral to the results.

\begin{table*}[h!]
\vspace{-0.5em}
\centering
\caption{Influence of increasing iteration count \textit{N} under With Constraint.}
\vspace{-0.5em}
\small
\setlength{\tabcolsep}{2pt} 
\resizebox{1\linewidth}{!}{
\begin{tabular*}{\textwidth}{@{\extracolsep{\fill}} c ccc ccc ccc}
\toprule
  &  \multicolumn{9}{c}{With Constraint}\\
  \cmidrule(lr){2-10}
    &
  \multicolumn{3}{c}{PREDCLS} & \multicolumn{3}{c}{SGCLS} & \multicolumn{3}{c}{SGDET} \\ 
  \cmidrule(lr){2-4} \cmidrule(lr){5-7} \cmidrule(lr){8-10} 
  &  mR@10  & mR@20  & mR@50 & mR@10  & mR@20  & mR@50  & mR@10 & mR@20  & mR@50 
\\
\midrule
\multicolumn{1}{l}{VISA}  & 46.9 &52.0 & 52.0&40.8 &42.5 &42.6  & 27.3 &27.3 & 30.7 \\

\multicolumn{1}{l}{VISA$_{N=2}$} & 47.9  & 53.0  & 53.0 & 41.7   & 43.4   & 43.4 & 28.7  & 28.7 & 31.9\\
\multicolumn{1}{l}{VISA$_{N=3}$}  & 48.1   & 53.3   & 53.3   & 42.2 & 44.0  & 44.0   & 28.9 & 28.9 &32.5 \\ 
\multicolumn{1}{l}{VISA$_{N=4}$}  & 48.9   & 53.8   & 53.8   & 42.8 & \textbf{44.5}  & \textbf{44.5}   & 29.5 & 29.5 &\textbf{33.1} \\ 
\multicolumn{1}{l}{VISA$_{N=5}$}  & \textbf{50.1}   & \textbf{53.9}   & \textbf{53.9}   & \textbf{43.0} & 44.3  & 44.3   & \textbf{29.6} & \textbf{29.6} &33.0 \\ 

\bottomrule
\end{tabular*}
}
\label{tab:supple_iters1}
\end{table*}

\begin{table*}[ht]
\centering
\caption{Influence of increasing iteration count \textit{N} under Semi Constraint.}
\vspace{-0.5em}
\small
\setlength{\tabcolsep}{2pt} 
\resizebox{1\linewidth}{!}{
\begin{tabular*}{\textwidth}{@{\extracolsep{\fill}} c ccc ccc ccc}
\toprule
  &  \multicolumn{9}{c}{Semi Constraint}\\
  \cmidrule(lr){2-10}
    &
  \multicolumn{3}{c}{PREDCLS} & \multicolumn{3}{c}{SGCLS} & \multicolumn{3}{c}{SGDET} \\ 
  \cmidrule(lr){2-4} \cmidrule(lr){5-7} \cmidrule(lr){8-10} 
  &  mR@10  & mR@20  & mR@50 & mR@10  & mR@20  & mR@50  & mR@10 & mR@20  & mR@50 
\\
\midrule
\multicolumn{1}{l}{VISA}  & 51.3 & 56.3  & 56.4  & 47.8  & 52.6 & 52.6 &  31.7 & 31.7& 33.2 \\
\multicolumn{1}{l}{VISA$_{N=2}$} & 52.5  & 56.9  & 57.0 & 48.2   & 53.6   & 53.6 & 31.9  & 31.9 & 32.7\\
\multicolumn{1}{l}{VISA$_{N=3}$}  & 52.8   & 57.1   & 57.2   & 48.5 & 53.9  & 53.9   & 32.0 & 32.0 &32.9 \\ 
\multicolumn{1}{l}{VISA$_{N=4}$}  & 53.0   & 57.3   & 57.3   & \textbf{48.6} & \textbf{54.0}  & \textbf{54.0}   & 32.1 & 32.1 &33.0 \\ 
\multicolumn{1}{l}{VISA$_{N=5}$}  & \textbf{53.1}   & \textbf{57.4}   & \textbf{57.4}   & 48.4 & 54.0  & 54.0   & \textbf{32.2} & \textbf{32.2} &\textbf{33.1} \\ 

\bottomrule
\end{tabular*}
}
\label{tab:supple_iters2}
\end{table*}

\begin{table*}[t!]
\centering
\caption{Influence of increasing iteration count \textit{N} under No Constraint.}
\vspace{-0.5em}
\small
\setlength{\tabcolsep}{2pt} 
\resizebox{1\linewidth}{!}{
\begin{tabular*}{\textwidth}{@{\extracolsep{\fill}} c ccc ccc ccc}
\toprule
  &  \multicolumn{9}{c}{No Constraint}\\
  \cmidrule(lr){2-10}
    &
  \multicolumn{3}{c}{PREDCLS} & \multicolumn{3}{c}{SGCLS} & \multicolumn{3}{c}{SGDET} \\ 
  \cmidrule(lr){2-4} \cmidrule(lr){5-7} \cmidrule(lr){8-10} 
  &  mR@10  & mR@20  & mR@50 & mR@10  & mR@20  & mR@50  & mR@10 & mR@20  & mR@50 
\\
\midrule
\multicolumn{1}{l}{VISA}&  65.8 &89.1 & 99.8 &52.0 &66.3  &71.4  & 30.7& 36.7& 50.4 \\
\multicolumn{1}{l}{VISA$_{N=2}$} & 66.0  & 90.8  & 99.8 & 53.1   & 67.0   & 72.0 & 31.8  & 37.2 & 51.4\\
\multicolumn{1}{l}{VISA$_{N=3}$}  & 66.2   & 91.2   & 99.9   & 53.5 & 67.2  & 72.5   & 32.0 & 37.5 &51.9 \\ 
\multicolumn{1}{l}{VISA$_{N=4}$}  & 66.3   & 91.3   & 99.9   & \textbf{53.7} & \textbf{67.5}  & \textbf{72.7}   & 32.2 & 37.7 &33.2 \\ 
\multicolumn{1}{l}{VISA$_{N=5}$}  & \textbf{66.3}   & \textbf{91.4}   & \textbf{99.6}   & 53.5 & 67.3  & 72.5   & \textbf{32.3} & \textbf{37.8} &\textbf{33.3} \\ 

\bottomrule
\end{tabular*}
}
\label{tab:supple_iters3}
\end{table*}

\subsection{Ablation Study of HSE}
In this section, we evaluated the effectiveness of the Hierarchical Semantics Extractor (HSE) by replacing it with a simple concatenation method. Specifically, the composite object feature $\mathbf{p}_{j,i}^{t}$ was concatenated with the integrated triplet embeddings $\mathbf{C}_{pre,(j,i)}^t$ and fed into the Spatial Encoder. The results of this ablation study were presented in Tables~\ref{tab:abla_HSE_with},~\ref{tab:abla_HSE_semi}, and~\ref{tab:abla_HSE_no}. The results demonstrate that using the concatenation approach led to a decrease of at least \textbf{-1.4\%} in mR@50 for the PREDCLS task under the No Constraint setting. This reduction was attributed to the simplistic visual-semantic fusion strategy, which failed to effectively integrate fine-grained visual and semantic features. The hierarchical structure of HSE, in contrast, facilitated a more sophisticated fusion process, capturing intricate relationships between visual and semantic information. This enhanced integration was crucial for mitigating biases and improving the accuracy of scene graph generation. The observed performance decline underscored the importance of maintaining hierarchical semantics extraction within the VISA framework to ensure unbiased VidSGG.

\subsection{Extended Study on the Influence of Iteration Count \textit{N}}

In this section, we investigated how varying the iteration number \textit{N} affects our framework's performance and determine the point at which computing costs outweigh performance gains. We incrementally increased the iteration count until this phenomenon occurs. The results of this analysis were detailed in Tables~\ref{tab:supple_iters1},~\ref{tab:supple_iters2}, and~\ref{tab:supple_iters3}. Examining the effect of iteration number $N$ on the performance of unbiased VidSGG, measured by mR@K, we observed that higher values of $N$ generally yield improved results. Notably, performance gains plateau at $N=4$, and by $N=5$, the computational costs begin to outweigh the benefits, even resulting in a decline in unbiased generation capabilities. We attributed this phenomenon to the limited training datasets, which caused the model's self-correction capabilities to reach a bottleneck. Consequently, future work may explore incorporating large language models (LLMs) to enhance the framework's adaptability and performance further. LLMs' inherent self-correction and language generation capabilities naturally complement this unbiased task. Expanding the training dataset could help overcome the current limitations, allowing for higher iteration counts without incurring prohibitive computational costs.

\vspace{-0.5em}
\section{Failure Cases}
\vspace{-0.5em}

\begin{figure}[t]
  \centering
  \includegraphics[width=0.48\textwidth]{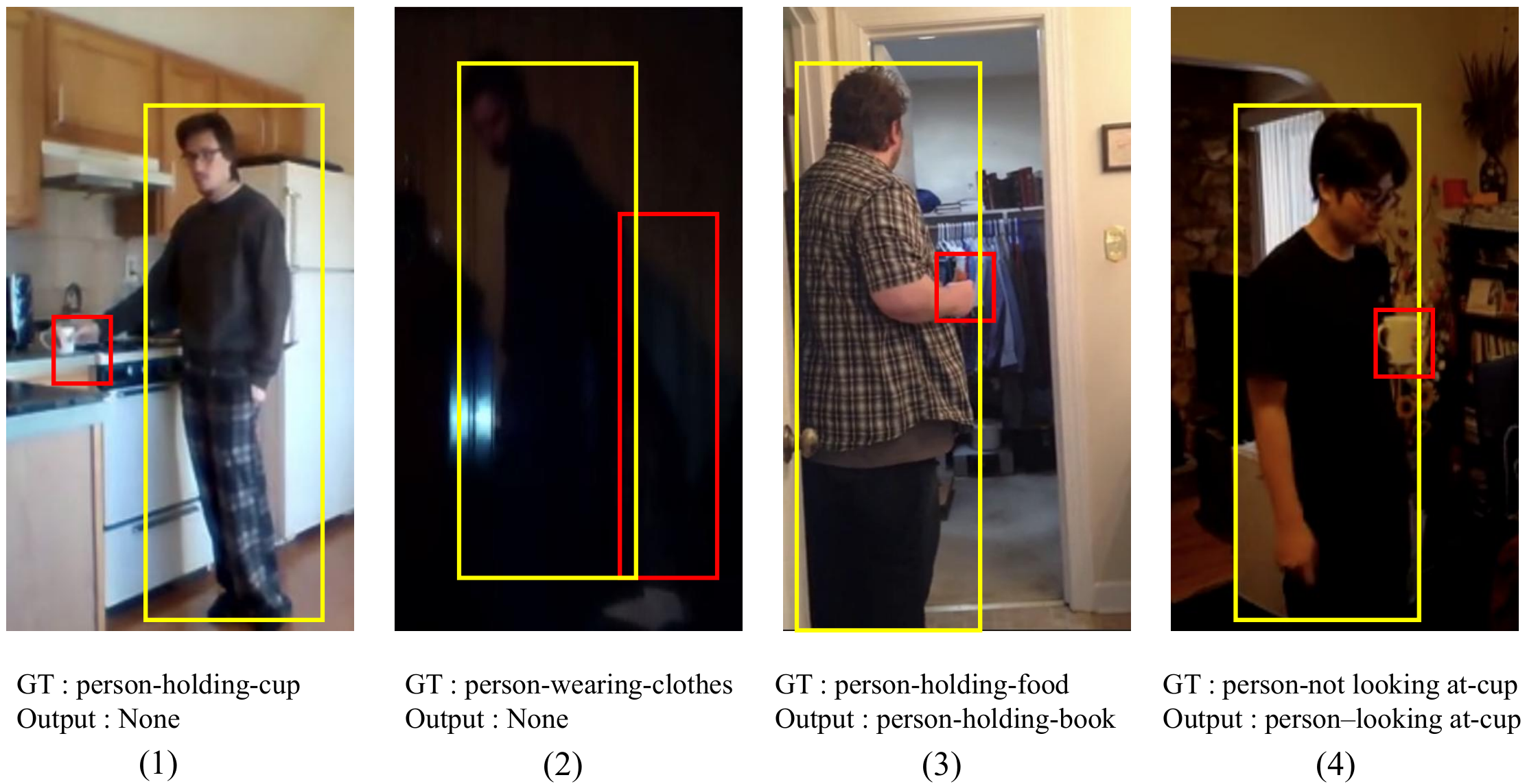}
  \caption{{\bf Prevalent Failure Cases.} (1) Undetected Small Objects (2) Noisy Annotations and Challenging Scenes (3) Ambiguity in Object Recognition (4) Ambiguity in Relationship Interpretation }
  \vspace{-0.5em}
  \label{fig:failure}
\end{figure}

To elucidate the limitations inherent in our model, we meticulously analyze the results and identify the most prevalent types of failure cases. We identify and illustrate several typical scenarios where VISA faces challenges, as depicted in Fig.~\ref{fig:failure}. (1) \textit{Undetected Small Objects.} Small objects like cups may be too diminutive for detection by the object detector, leading to the omission of related triplets in VISA's output. (2) \textit{Noisy Annotations and Challenging Scenes in AG.} This includes incorrect annotations, low-resolution videos, and extreme scene conditions. For instance, scenes that are too dim to discern events accurately. (3) \textit{Ambiguity in Object Recognition.} Certain objects are indistinguishable even to human observers, such as differentiating between a person holding a book and food. (4) \textit{Ambiguity in Relationship Interpretation.} Some relationships are also challenging to discern, like determining whether a person is looking at a cup or not.

We speculate that Failure (1), the adoption of a more advanced object detector could potentially offer a solution. Addressing Failure (2) may involve comprehensive data cleansing efforts. As for Failures (3) and (4), which we attribute to the intrinsic constraints of human-labeled annotations, an unsupervised learning approach might present a viable resolution.

\section{Supplementary visualization results}
\begin{figure}[h]
  \centering
  \includegraphics[width=0.5\textwidth]{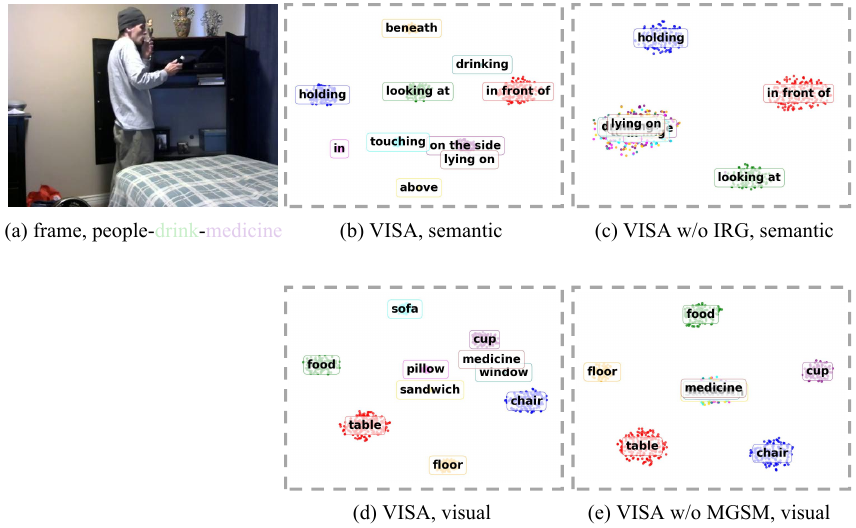}
  \vspace{-1em}
  \caption{{\bf Supplementary visualizations.}}
  \vspace{-1em}
  \label{fig:supple_visual}
\end{figure}

Figure~\ref{fig:supple_visual} shows our t-SNE results for semantic (b,c) and visual (d,e) features, effectively separating high- and low-frequency classes (e.g., drink, medicine) and mitigating both visual and semantic biases. This figure also offers more results of complex scenes, featuring low-frequency predicates (e.g., \emph{drink}) and nouns (e.g., \emph{medicine}) that are smaller and harder to recognize.

\section{Recall@K explanation}Recall@K(R@K) was used as a standard metric for previous VidSGG method~\cite{nag2023unbiased,cong2023reltr}. However, we excluded R@K in the main paper due to reporting bias identified by leading unbiased image-based SGG methods~\cite{misra2016seeing,yan2020pcpl,tang2020unbiased}. R@K favors high-frequency representations, a bias overlooked in previous VidSGG methods. The R@K results were obtained under the same experimental settings as in the main paper (Table~\ref{tab:R@K_sota}) and are included for completeness, though they are not the primary focus in VidSGG.

\begin{table}[h]
\centering
\vspace{-0.5em}
\caption{Quantitative R@K results.}
\vspace{-0.5em}
\large
\setlength{\tabcolsep}{2pt} 
\renewcommand{\arraystretch}{0.9} 
\resizebox{\columnwidth}{!}{
\begin{tabular}{l l c c c c c c c c c}
\toprule
\multirow{2}{*}{Constraint} & \multirow{2}{*}{Method} & 
\multicolumn{3}{c}{PREDCLS} & 
\multicolumn{3}{c}{SGCLS} &  
\multicolumn{3}{c}{SGDET}  \\ 
\cmidrule(lr){3-5} \cmidrule(lr){6-8} \cmidrule(lr){9-11} 
& & R@10 & R@20 & R@50 & R@10 & R@20 & R@50 & R@10 & R@20 & R@50 \\
\midrule

\multirow{3}{*}{\rotatebox{90}{With}} 
& TEMPURA & 68.8 & 71.5 & 71.5 & 47.2 & 48.3 & 48.3 & 28.1 & 33.4 & 34.9 \\
& FloCoDe & 70.1 & 74.2 & 74.2 & 48.4 & 51.2 & 51.2 & 31.5 & 38.4 & 42.4 \\
& \textbf{VISA (Ours)} & \cellcolor{lightblue}\textbf{70.2} & \cellcolor{lightblue}\textbf{74.9} & \cellcolor{lightblue}\textbf{75.3} & \cellcolor{lightblue}\textbf{49.1} & \cellcolor{lightblue}\textbf{51.9} & \cellcolor{lightblue}\textbf{52.3} & \cellcolor{lightblue}\textbf{32.2} & \cellcolor{lightblue}\textbf{39.3} & \cellcolor{lightblue}\textbf{43.8} \\
\midrule

\multirow{2}{*}{\rotatebox{90}{Semi}} 
& TEMPURA & 66.9 & 69.6 & 69.7 & 48.3 & 50.0 & 50.0 & 28.1 & 33.3 & 34.8 \\
& \textbf{VISA (Ours)} & \cellcolor{lightblue}\textbf{70.8} & \cellcolor{lightblue}\textbf{76.6} & \cellcolor{lightblue}\textbf{76.7} & \cellcolor{lightblue}\textbf{56.8} & \cellcolor{lightblue}\textbf{61.2} & \cellcolor{lightblue}\textbf{61.2} & \cellcolor{lightblue}\textbf{35.9} & \cellcolor{lightblue}\textbf{40.9} & \cellcolor{lightblue}\textbf{42.0} \\
\midrule

\multirow{3}{*}{\rotatebox{90}{No}} 
& TEMPURA & 80.4 & 94.2 & 99.4 & 56.3 & 64.7 & 67.9 & 29.8 & 38.1 & 46.4 \\
& FloCoDe& 82.8 & 97.2 & \textbf{99.9} & 57.4 & 66.2 & 68.8 & 32.6 & 43.9 & 51.6 \\
& \textbf{VISA (Ours)} & \cellcolor{lightblue}\textbf{83.5} & \cellcolor{lightblue}\textbf{98.5} & \cellcolor{lightblue}\textbf{99.9} & \cellcolor{lightblue}\textbf{58.0} & \cellcolor{lightblue}\textbf{67.2} & \cellcolor{lightblue}\textbf{70.1} & \cellcolor{lightblue}\textbf{33.2} & \cellcolor{lightblue}\textbf{44.7} & \cellcolor{lightblue}\textbf{52.4} \\
\bottomrule
\end{tabular}
}
\label{tab:R@K_sota}
\vspace{-0.5em}
\end{table}

As shown in Table~\ref{tab:R@K_sota}, VISA surpasses all previous methods across all R@K metrics in unbiased VidSGG.

\section{More details on PVSG and 4DPVSG}We keep the baseline unchanged from the original paper~\cite{yang2023panoptic,yang2023psg4d}. For PVSG, we adopt a Mask2Former-based method for image panoptic segmentation, then use UniTrack for visual representations, and finally apply a Transformer encoder for relationship prediction. For 4DPVSG, we process 3D video clips with Mask2Former for frame-level panoptic segmentation, link instance embeddings across frames via UniTrack, and employ a Spatial-Temporal Transformer to incorporate temporal context and inter-object interactions. 
{
    \small
    \bibliographystyle{ieeenat_fullname}
    \bibliography{main}
}


\end{document}